\pdfoutput=1

\documentclass[11pt]{article}

\usepackage[final]{acl}

\usepackage{times}
\usepackage{latexsym}

\usepackage[T1]{fontenc}

\usepackage[utf8]{inputenc}

\usepackage{microtype}

\usepackage{inconsolata}

\usepackage{graphicx}

\usepackage{multirow}
\usepackage{booktabs}
\usepackage{graphicx}
\usepackage{subcaption}
\usepackage{amssymb}
\usepackage{CJKutf8}
\usepackage{amsmath} 
\usepackage{hyperref}
\usepackage{enumitem}
\usepackage{color}
\graphicspath{{images/}}
\usepackage{algorithm}
\usepackage{algpseudocode}
\usepackage{tgadventor}
\usepackage{tikz}
\usepackage{xcolor}

\newcommand{\zh}[1]{\protect\begin{CJK*}{UTF8}{gbsn}#1\protect\end{CJK*}}

\definecolor{german}{HTML}{FFCC67}
\definecolor{english}{HTML}{38FFF8}

\title{XRAG: Cross-lingual Retrieval-Augmented Generation}

\author{Wei Liu$^{1}$\thanks{Work done during an internship at Amazon.}, Sony Trenous$^{2}$, Leonardo F. R. Ribeiro$^{2}$, Bill Byrne$^{2}$, Felix Hieber$^{2}$ \\ $^{1}$Heidelberg Institute for Theoretical Studies gGmbH  \\ $^{2}$Amazon AGI \\ \texttt{wei.liu@h-its.org}, \texttt{\{trenous,leonribe,willbyrn,fhieber\}@amazon.com}}

\begin{document}
\maketitle

\begin{abstract}
We propose XRAG, a novel benchmark design\-ed- to evaluate the generation abilities of LLMs in cross-lingual Retrieval-Augmented Generation (RAG) settings where the user language does not match the retrieval results. XRAG is constructed from recent news articles to ensure that its questions require external know\-ledge to be answered. It covers the real-world scenarios of monolingual and multilingual retrieval, and provides relevancy annotations for each retrieved document. Our novel dataset construction pipeline results in questions that require complex reasoning, as evidenced by the significant gap between human and LLM performance. Consequently, XRAG serves as a valuable benchmark for studying LLM reasoning abilities, even before considering the additional cross-lingual complexity. Experimental results on five LLMs uncover two previously unreported challenges in cross-lingual RAG: 1) in the monolingual retrieval setting, all evaluated models struggle with response language correctness; 2) in the multilingual retrieval setting, the main challenge lies in reasoning over retrieved information across languages rather than generation of non-English text.


\end{abstract}
\section{Introduction}
\begin{figure}[t]
\centering
\includegraphics[scale=0.425,trim=0 0 0 0]{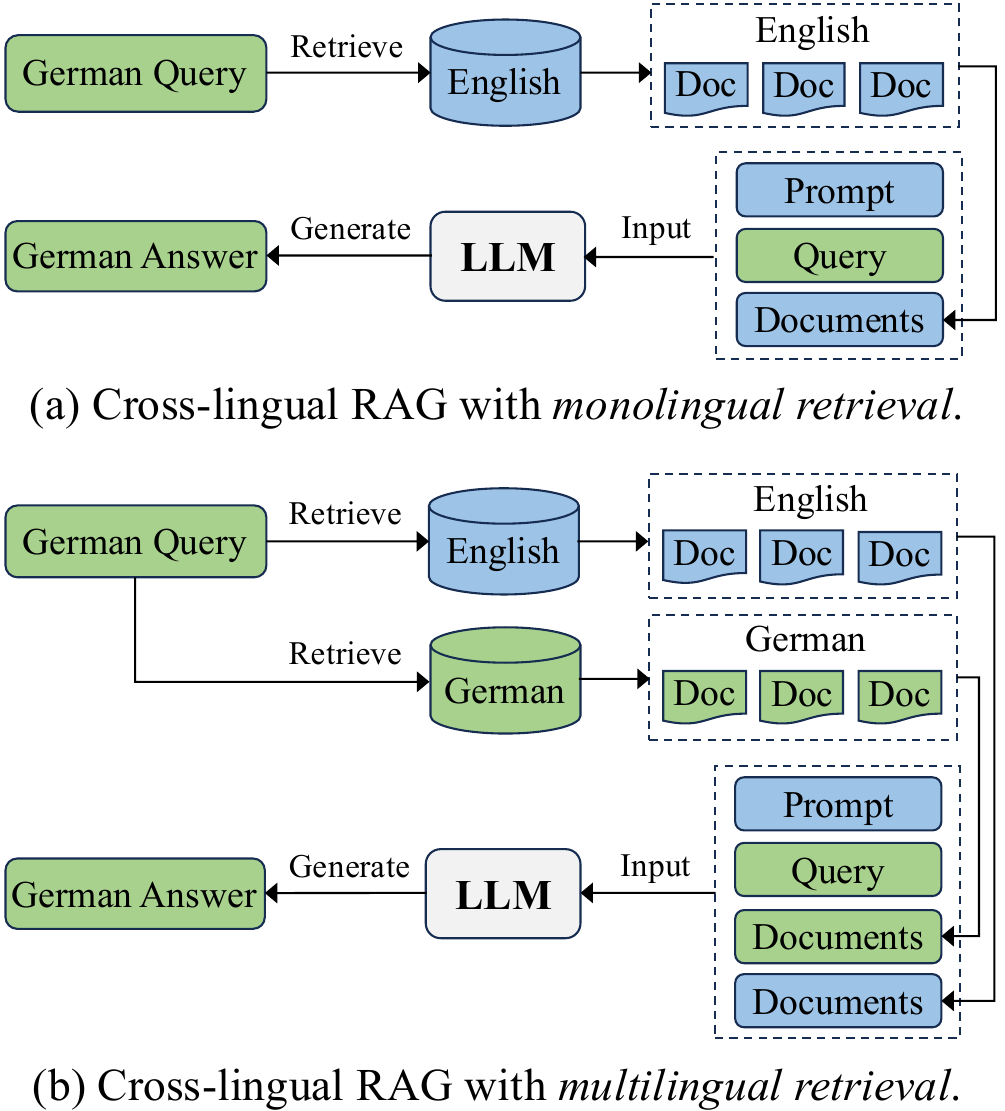}
\setlength{\abovecaptionskip}{6pt}
\setlength{\belowcaptionskip}{-16pt}
\caption{Two cases of cross-lingual RAG: (a) monolingual retrieval, where the LLM uses  retrieved English documents to respond to a German query; (b) multilingual retrieval, where the LLM uses  retrieved English and German documents to respond to a German query.}
\label{fig:motivation}
\end{figure}

Retrieval-augmented generation (RAG) augments large language models (LLMs) by retrieval of relevant documents with the aim of improving response quality~\citep{rag}. The widespread adoption of RAG has prompted many recent studies to evaluate specific capabilities of LLMs in RAG settings, such as robustness to noise \citep{domainrag}, information integration~\citep{chen2024benchmarking}, time sensitivity~\citep{kasai2023realtime}, multi-hop reasoning~\citep{tang2024multihoprag} and conversational QA~\cite{roy-etal-2024-learning}. Notably, these evaluations are in monolingual settings
in which questions and retrieved documents are in the same language.

Real-world deployments of RAG systems also need to handle cross-lingual use cases, where the user's language does not match that of the retrieved documents. The simplest scenario is \textbf{Cross-lingual RAG with Monolingual Retrieval}~\cite{asai2023selfrag}, where users in multiple locales are served by a single RAG system that accesses an English-only knowledge base, as illustrated in Figure \ref{fig:motivation}a. This setup applies to, for example, a general-purpose RAG system that relies solely on English web search or a corporate helpdesk with an internal database available only in English. A more complex scenario is \textbf{Cross-lingual RAG with Multilingual Retrieval}, where RAG systems combine information from both English and the user's language to generate a response (see Figure \ref{fig:motivation}b). This is a common situation in that native-language sources often contain culturally or geographically specific knowledge, with English resources providing additional, more general information.\footnote{An initial study on a proprietary dataset of real-world LLM traffic from non-English users in Germany, Japan, and Spain found that using only English or native-language search results was inferior to combining both (see Appendix \ref{app:sony}).}

\vspace{-1pt}
Due to the absence of relevant benchmarks, we lack an understanding of how well LLMs can handle such cross-lingual RAG scenarios. A potential solution is to use existing cross-lingual open-domain question-answering datasets, such as XQA~\citep{liu-etal-2019-xqa} and XOR QA~\citep{asai-etal-2021-xor}, for evaluation~\cite{chirkova-etal-2024-retrieval}. Yet these datasets only cover limited cross-lingual scenarios; in particular, the documents used to answer questions are  in English, which hinders the evaluation of LLMs in more complex multilingual scenarios (i.e., Figure \ref{fig:motivation}b). Moreover, the questions in these datasets tend to be relatively simple (e.g. span extraction questions) and often can be answered without retrieval.\footnote{~\citet{chirkova-etal-2024-retrieval} shows that 47.5\% of questions in XORQA can be answered by Command-R without retrieval.} Due to these shortcomings, these datasets do not measure the true cross-lingual capabilities of LLMs in RAG settings.

\vspace{-1pt}
To address this gap, we introduce XRAG, a benchmark for evaluating the Question Answering capabilities of LLMs in cross-lingual RAG scenarios, where some information must be extracted from retrieved documents that are not in the user's language. The benchmark features natural-sounding questions that require cross-document reasoning and are \textbf{challenging for LLMs even in an English monolingual RAG setting} (GPT-4o achieves only 62.4\%  accuracy, see Table~\ref{table:mono}). We develop a novel LLM-based question generation workflow using recent news articles, ensuring that current frontier models are \textbf{unable to answer the questions without retrieval} (GPT-4o accuracy is 6.3\% without retrieval, see Table \ref{table:property}). To guarantee a high-quality dataset, we employ extensive human Quality Assurance, resulting in  \textbf{few  ambiguous or noisy questions} (under 8\%, see Section ~\ref{sec:corpus_ana}).  In addition to English, the benchmark spans \textbf{four widely spoken and linguistically diverse languages} (Arabic, Chinese, German, and Spanish). 

XRAG comprises two sub-tasks, corresponding to the \textit{monolingual retrieval} and the \textit{multilingual retrieval} settings of cross-lingual RAG. For each non-English language, we provide a directly comparable English monolingual RAG baseline task. Each instance in the XRAG benchmark consists of a question, a gold answer, two supporting articles that together answer the question, and six topically related but non-answering distracting articles. This allows us to approximate  realistic RAG settings with imperfect retrieval in evaluating the Question Answering abilities of LLMs.

We evaluate five LLMs, including both closed- and open-source models, on XRAG. In summary, our contributions are:
\vspace{-2pt}
\begin{itemize}
\setlength{\itemsep}{-0.3em}  
  \setlength{\topsep}{-0.3em}   
  \item[(1)] We introduce XRAG, a novel benchmark designed to evaluate the performance of LLMs in two cross-lingual RAG scenarios.  
  \item[(2)] We propose a novel method for generating challenging cross-document QA pairs from News Crawl, resulting in natural questions that current LLMs cannot answer using only their parametric knowledge.
  \item[(3)]  We find that in the \textit{monolingual retrieval} setting, all evaluated LLMs face issues with Response Language Correctness-an issue that has received little attention from the research community.
  \item[(4)] In the \textit{multilingual retrieval} setting, the primary challenge for LLMs does not lie in non-English generation, but in reasoning over retrieved information across languages.
\end{itemize}
\vspace{-2pt}

\vspace{-4pt}
\section{Related Work}
\vspace{-2pt}

\begin{table*}[t]
\centering
\Large
\scalebox{0.44}{
\renewcommand{\arraystretch}{1.4}
\begin{tabular}{l|l|l|l}
\hline
RAG Setting                                                                                                    & Field                                  & Language                        & Content                                                                                                                                                                                                                                                                                                            \\ \hline
\multicolumn{1}{l|}{}                                                                                   & Question                               & \colorbox{german}{German}  & Wie viel haben Walmart und ALDI zusammen für die Opfer des Hurrikans Helene 2024 gespendet?                                                                                                                                                                                                                        \\ \cline{2-4} 
\multicolumn{1}{l|}{}                                                                                   & Answer                                 & \colorbox{german}{German}  & Die gesamten Spenden überstiegen 11 Millionen Dollar.                                                                                                                                                                                                                                                              \\ \cline{2-4} 
\multicolumn{1}{l|}{}                                                                                   &                                        & \colorbox{english}{English} & Walmart, Sam’s Club and the Walmart Foundation are increasing their commitment to \$10 million to Hurricane Helene Relief Effort...                                                                                                                                                                                \\ \cline{3-4} 
\multicolumn{1}{l|}{}                                                                                   & \multirow{-2}{*}{\begin{tabular}[c]{@{}l@{}}Supporting\\ Articles\end{tabular}}  & \colorbox{english}{English} & The American Red Cross recognizes ALDIfor its pledge of \$1,000,000. By making a donation to Hurricane Helene Relief...                                                                                                                                                                                  \\ \cline{2-4} 
\multicolumn{1}{l|}{}                                                                                   &                                        & \colorbox{english}{English} & Walmart Canada reaches new giving milestone of \$750 million raised and donated to charities and non-profits across Canada...                                                                                                                                                                                      \\ \cline{3-4} 
\multicolumn{1}{l|}{\multirow{-6}{*}{\begin{tabular}[c]{@{}l@{}}Monolingual\\ Retrieval\end{tabular}}}  & \multirow{-2}{*}{\begin{tabular}[c]{@{}l@{}}Distracting\\ Articles\end{tabular}} & \colorbox{english}{English} & Aldi has donated £2,000 to charities in Gloucestershire to help support those in need during the school holidays. The donations...                                                                                                                                                                                 \\ \hline
\multicolumn{1}{l|}{}                                                                                   & Question                               & \colorbox{german}{German}  & Welches Land gewann seine erste Goldmedaille bei den Olympischen Spielen 2024 früher, die Vereinigten Staaten oder Deutschland?                                                                                                                                                                                    \\ \cline{2-4} 
\multicolumn{1}{l|}{}                                                                                   & Answer                                 & \colorbox{german}{German}  & \begin{tabular}[c]{@{}l@{}}Deutschland. Beide Länder gewannen am 27. Juli bei den Schwimmwettbewerben ihre ersten Goldmedaillen, aber die USA gewannen\vspace{-6pt}\\ ihre Medaille erst im letzten Wettkampf des Tages – später als Deutschland.\end{tabular} \\ \cline{2-4} 
\multicolumn{1}{l|}{}                                                                                   &                                        & \colorbox{german}{German}  & Lukas Märtens hat am 27. Juli in Paris den olympischen Titel über 400 m Freistil gewonnen. Der Magdeburger siegte in 3:41,78...                                                                                                                                                                                    \\ \cline{3-4} 
\multicolumn{1}{l|}{}                                                                                   & \multirow{-2}{*}{\begin{tabular}[c]{@{}l@{}}Supporting\\ Articles\end{tabular}}  & \colorbox{english}{English} & In the last swimming race of July 27, the U.S. took its first gold medal of the 2024 Olympics, winning the 4×100-meter freestyle...                                                                                                                                                                                \\ \cline{2-4} 
\multicolumn{1}{l|}{}                                                                                   &                                        & \colorbox{german}{German}  & Im Rahmen von noch bevorstehenden Qualifikationsevents können sich weitere Sportler noch für die Spiele in Paris qualifizieren...                                                                                                                                                                                  \\ \cline{3-4} 
\multicolumn{1}{l|}{\multirow{-6}{*}{\begin{tabular}[c]{@{}l@{}}Multilingual\\ Retrieval\end{tabular}}} & \multirow{-2}{*}{\begin{tabular}[c]{@{}l@{}}Distracting\\ Articles\end{tabular}} & \colorbox{english}{English} & The United States Olympic \& Paralympic Committee have announced the 592-member 2024 U.S. Olympic team ready to compete...                                                                                                                                                                                         \\ \hline
\end{tabular}}
\setlength{\abovecaptionskip}{6pt}
\setlength{\belowcaptionskip}{-12pt}
\caption{Two instances from XRAG, each consisting of a question, a gold answer, two supporting articles, and six distracting articles (two are shown). 
In the \textit{monolingual retrieval} setting,  all supporting and distracting articles are in English;
in the \textit{multilingual retrieval} setting, the supporting and distracting articles are in the question language and in English. LLMs should answer these questions based on the supporting articles while ignoring the distractors.
}
\label{table:example_ml}
\end{table*}

There are extensive recent investigations into characterizing the Question Answering capabilities of LLMs in RAG settings. \citet{vu-etal-2024-freshllms} construct a a dynamic QA benchmark, FreshQA, that tests the ability of LLMs to use up-to-date world knowledge to solve questions.~\citet{rgb} introduce RGB as a benchmark to analyze fundamental abilities of LLMs in RAG systems, such as noise robustness and negative rejection. \citet{tang2024multihoprag} propose MultiHop-RAG, which focuses on whether retrieval-enhanced LLMs can retrieve and reason over multiple pieces of supporting evidence. The Comprehensive RAG Benchmark, created by ~\citet{crag}, aims to assess whether LLMs are able to answer different types of questions, ranging from simple to complex.~\citet{MIRAGE-Bench} present MIRAGE-Bench, a multilingual RAG benchmark constructed from Wikipedia, to evaluate RAG systems performance in different languages. As noted, these are in monolingual settings, whereas we aim to benchmark LLMs performance in cross-lingual RAG scenarios.~\citet{chirkova-etal-2024-retrieval} is one of few studies that evaluate the cross-lingual capabilities of LLMs in RAG systems. They conduct an analysis of existing cross-lingual open-domain question-answering datasets~\citep{asai-etal-2021-xor}.  Motivated by this prior work, XRAG is a new cross-lingual benchmark that covers a wider range of scenarios and consists of questions designed to require external knowledge to answer, thereby providing a more accurate reflection of the cross-lingual capabilities of LLMs in RAG.

\vspace{-3pt}
Evaluation of cross-lingual NLP systems is a long-standing research problem. Relatively recent work has focused on performance in specific NLP tasks such as NLI~\citep{conneau-etal-2018-xnli}, summarization~\citep{wang-etal-2022-survey}, retrieval question answering~\citep{roy-etal-2020-lareqa}, and open-domain question answering~\citep{naacl-2021-main}. With the advent of large language models, cross-lingual evaluation has expanded to include few-shot or even zero-shot settings. \citet{wang-etal-2023-zero} investigate GPT-4 performance for cross-lingual summarization in a zero-shot setting and find that it performs competitively with  finetuned mBART-50. ~\citet{ahuja-etal-2023-mega} further evaluate the performance of generative models on 15 tasks, covering classification, sequence labeling, and generation. 

\vspace{-3pt}
We note that our focus is on retrieval augmented generation from multilingual document retrieval, and not on  the document  retrieval task itself. However in Sections ~\ref{sec:find_pairs} and \ref{sec:pre_retrieved} we discuss how we use monolingual and multilingual dense document retrieval techniques in constructing XRAG.  Our work aligns with recent efforts in evaluating cross-lingual performance of LLMs, with a focus on retrieval augmented generation and cross-lingual answer generation in particular.

\vspace{-2pt}
\section{XRAG - Cross-lingual RAG Benchmark}
\vspace{-2pt}
We define the task of cross-lingual RAG as follows: given a question $q$, the LLM is prompted to generate an answer $\tilde{a}$ in the same language as the question by referring to a collection of $m$ articles ${\rm D}=\{d_1, d_2, ..., d_m\}$ that contains articles in a language different than the question:
\vspace{-5pt}
\begin{equation}
    \centering
    \begin{aligned}
        & \tilde{a} \leftarrow {\rm{\tt LLM}}(q, {\rm D}, {\sf prompt})  \\
        & {\tt Language}(q) = {\tt Language}(\tilde{a}) \\
        & \exists\,d_i \in {\rm D},\; {\tt Language}(d_i) \neq {\tt Language}(q)
    \end{aligned}
    \vspace{-4pt}
\end{equation}
Figure \ref{fig:motivation} shows two cases, in which LLMs need to use information from \textit{English} articles to generate \textit{German} responses to \textit{German} questions. The goal of our benchmark is to enable an understanding of \textbf{how well LLMs perform generation in such cross-lingual RAG scenarios}.

\begin{figure*}[t]
\centering
\includegraphics[scale=0.412,trim=0 0 0 0]{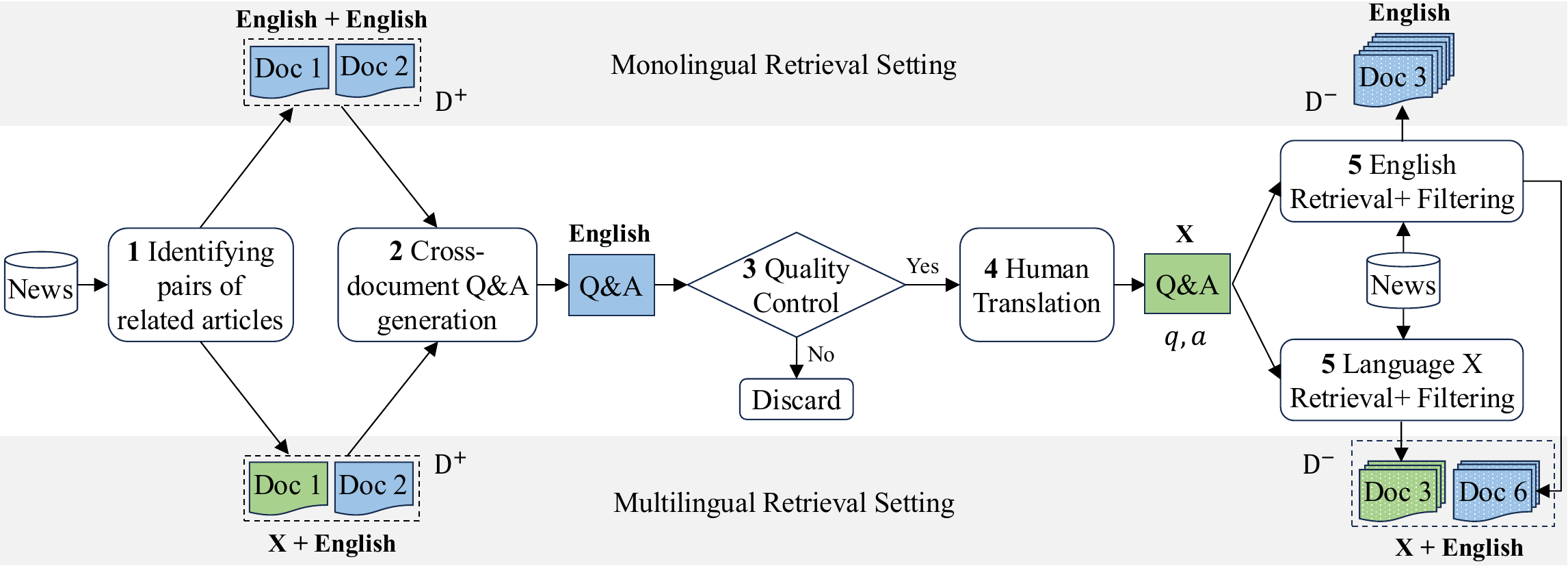}
\setlength{\abovecaptionskip}{-4pt}
\setlength{\belowcaptionskip}{-10pt}
\caption{Each instance $(q, a, {\rm D}^+, {\rm D}^-)$ in XRAG — where $q$ is the question, $a$ the gold answer, $\rm D^+$ the supporting articles, and $\rm D^-$ the distractors — is constructed as follows: (1) find two related articles; (2) generate an English cross-document Q\&A pair using the two articles; (3) evaluate the quality of the Q\&A pair; (4) translate the Q\&A pair into language X $\in$ \{German, Spanish, Chinese, Arabic\}; and (5) collect distracting articles for the question.}
\label{fig:construction}
\end{figure*}

Each instance $(q, a, {\rm D}^+, {\rm D}^-)$ in XRAG consists of a question $q$, a golden answer $a$, two supporting articles ${\rm D}^+$, and several distracting articles ${\rm D}^-$.  The supporting articles each contribute partial information needed to answer the question, and only together do they provide a complete answer; in contrast, the distracting articles are topically related but cannot answer the question. Taken together, this simulates a realistic RAG scenario with imperfect retrieval, where we can control the quality of the grounding by the inclusion of distractors. Questions in XRAG are cross-document questions, requiring reasoning across the two supporting articles to answer, while ignoring the distracting articles. Our benchmark considers two real-world~cross-lingual RAG scenarios: the \textit{monolingual retrieval} scenario and the \textit{multilingual retrieval} scenario.

In the \textit{monolingual retrieval} setting, LLMs rely on English articles to generate an answer. This occurs when users in multiple locales are served by a single cross-lingual RAG system that has access only to an English knowledge base. In this paper, we consider questions in four languages: German (de), Spanish (es), Chinese (zh), and Arabic (ar):
\vspace{-5pt}
\begin{equation}
    \begin{aligned}
         & {\tt Language}(q) \in \{\text{de}, \text{es}, \text{zh}, \text{ar}\} \\
         & {\tt Language}({\rm D}^+) = {\tt Language}({\rm D}^-) = \text{en}
    \end{aligned}
    \vspace{-4pt}
\end{equation}
These four are widely used in the research community~\citep{macko-etal-2023-multitude} and represent a range of cross-lingual challenges, ranging from easy (es-en) to challenging (zh-en)~\citep{yang-etal-2022-crop}.

In a \textit{multilingual retrieval} setting, LLMs use articles in both the question language and other languages to answer a question. This corresponds to a cross-lingual RAG scenario where documents in a resource-rich language provide additional information for LLMs to answer questions in a second language. Similarly, we consider four languages:
\vspace{-4pt}
\begin{equation}
    \begin{aligned}
         & {\tt Language}(q) \in \{\text{de}, \text{es}, \text{zh}, \text{ar}\} \\
         & {\tt Language}({\rm D}^+) = \{\text{en}, {\tt Language}(q) \} \\
         & {\tt Language}({\rm D}^-) = \{\text{en}, {\tt Language}(q) \}
    \end{aligned}
    \vspace{-4pt}
\end{equation}
Table \ref{table:example_ml} gives examples of \textit{monolingual retrieval} and \textit{multilingual retrieval} from XRAG.

\begin{figure*}[t]
\centering\includegraphics[scale=0.41,trim=0 0 0 0]{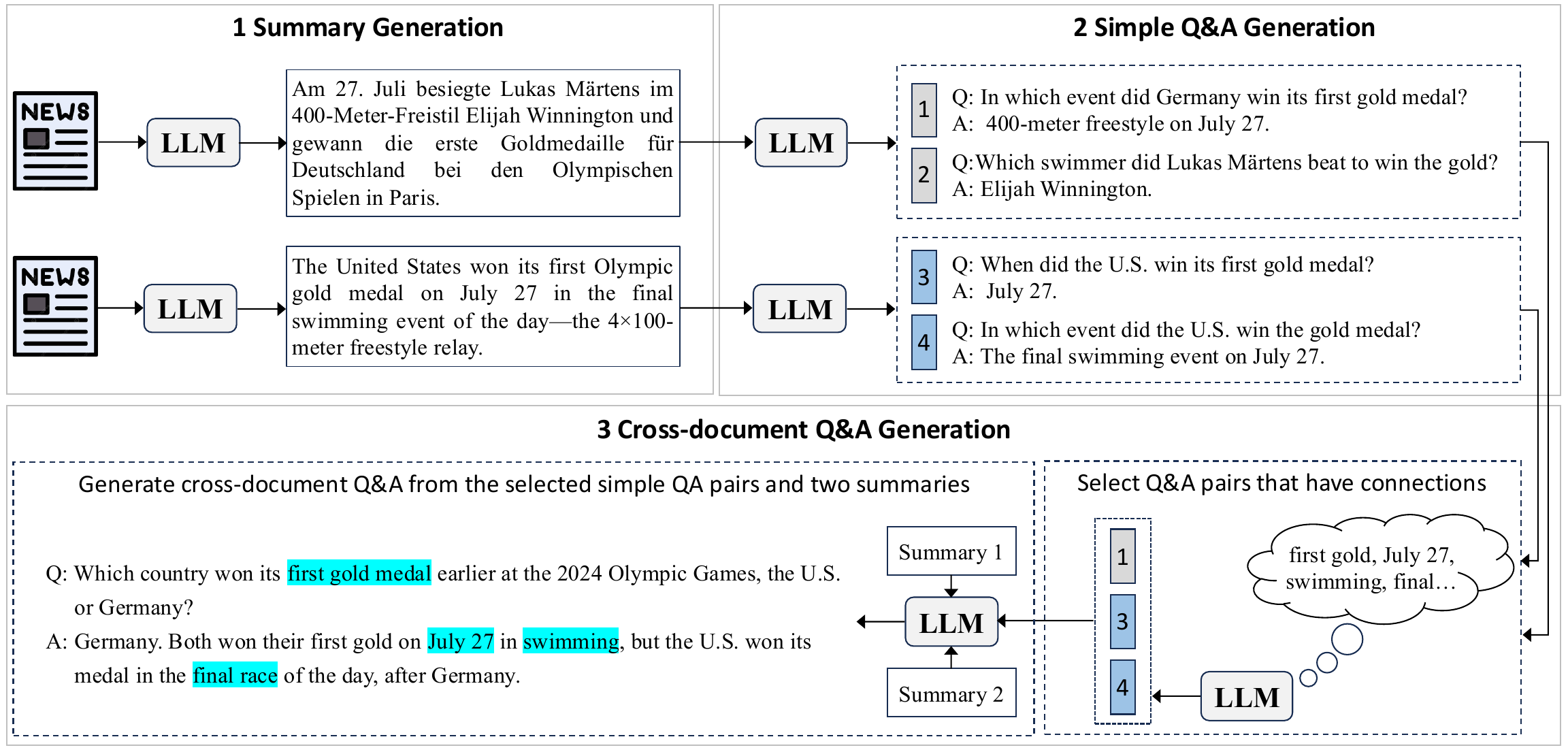}
\setlength{\abovecaptionskip}{-6pt}
\setlength{\belowcaptionskip}{-10pt}
\caption{LLM-based workflow for generating English cross-document questions from a pair of related articles: (1) generate a summary for each article; (2) create simple English Q\&A pairs from each summary that require only one-step reasoning; (3) identify connections between the two sets of Q\&A pairs, select related ones, and construct a new Q\&A pair that requires reasoning across multiple pieces of information from the selected pairs and summaries.}
\label{fig:workflow}
\end{figure*}
\vspace{-2pt}
\section{XRAG Construction}
\vspace{-1pt}
Figure \ref{fig:construction} shows the overall XRAG construction process.
We begin with English, German, Spanish, Chinese, and Arabic news articles from News Crawl between June 1, 2024, and November 30, 2024. This timeframe ensures that the articles are dated after the knowledge cutoff of LLMs such as GPT-4 and Claude 3.5. Questions created from these articles are more likely to require external knowledge to answer. 

\vspace{-2pt}
\subsection{Identifying pairs of related articles}
\label{sec:find_pairs}
\vspace{-1pt}
To generate natural cross-document questions from a pair of articles, the articles must be topically related; otherwise, the generated questions may seem artificial~\citep{welbl-etal-2018-constructing}. 

For the \textit{monolingual retrieval} setting, 
we construct a bipartite graph linking English articles with the entities in their titles. We then use depth-first search to find pairs of articles that share at least two entities in their titles. These article pairs serve as related English articles for generating cross-document questions.  

For the \textit{multilingual retrieval} setting, we use international events from Wiki 2024 and a multilingual dense retriever to search across different languages for articles related to the events. We then group articles in different languages about the same event  to form related article pairs. 

We provide a more detailed explanation of how to locate relevant articles in English or across languages in Appendices \ref{app:find_pairs}.

\vspace{-3pt}
\subsection{English cross-document Q\&A generation}
\label{sec:q_generation}
\vspace{-1pt}
We design an LLM-based workflow\footnote{We use GPT-4o-2024-08-06.} to generate\- natural and coherent English cross-document questions from news articles. Figure \ref{fig:workflow} shows an~overview of the generation workflow. 

\noindent \textbf{Step 1: Summary Generation}. Given a pair of related articles either in English or in English and another language, we prompt the LLM to create a summary for each article that (1) is accurate and concise; (2) covers the key points; and (3) has little lexical overlap with the article (the prompt is shown in Figure \ref{prompt:summary}). These summaries are then used to generate questions in Step 2. There are two reasons for this: generating questions\- directly from articles often leads to questions with high overlap in wording, making it easy to answer through string matching; and direct question generation from articles can focus on trivial details, whereas generating questions from summaries tends to produce questions about the main points of the articles. 

\noindent \textbf{Step 2: Simple Q\&A Generation}. Our goal is to create cross-document questions that require information from two articles to answer. However, we find that generating such questions in one step is difficult. LLMs often create questions that simply link two separate questions with "and". Instead, we first prompt the LLM to generate simple English Q\&A pairs from each summary that can be answered with one step of reasoning (the prompt is shown in Figure \ref{prompt:simple_qa}). For example, from a German report about Germany's first gold medal at the 2024 Olympics, the LLM generates Q\&A pairs like: (\textbf{q}: In which event did Germany win its first gold medal? \textbf{a}: 400-meter freestyle on July 27).

\noindent \textbf{Step 3: Cross-document Q\&A Generation}. After generating simple Q\&A pairs from two summaries, we prompt the LLM to: (1) identify connections between the two sets of Q\&A pairs; (2) select related Q\&A pairs from the two sets, ensuring that at least one pair is chosen from each source; and (3) formulate new questions that require reasoning across multiple pieces of information drawn from the selected Q\&A pairs. Since the selected Q\&A pairs originate from different source articles, answering the newly generated questions necessitates integrating information from both sources, thus resulting in cross-document questions. For example, using the simple Q\&A pairs in Figure \ref{fig:workflow}, the LLM finds links such as "first gold medal", "swimming race", "final" and the date "July 27" between the two sets of simple Q\&A pairs. The LLM then generates a comparison question: "Which country won its first gold medal earlier at the 2024 Olympic Games, the U.S. or Germany?". We then ask the LLM to generate an answer to the question using information from the selected simple Q\&A pairs and the two summaries (the prompt for answer generation is in Figure \ref{prompt:crossdoc_a}). Inspired by \citet{crag}, we focus on four types of cross-document questions: aggregation, comparison, multi-hop, and set questions. We present the definition of the four types of questions in Table \ref{table:q_types}, and the prompts to generate these questions in Figures \ref{prompt:aggregation}, \ref{prompt:comparison}, \ref{prompt:multihop}, and \ref{prompt:set}.

\vspace{-4pt}
\subsection{Quality Control and Human Translation}
\label{sec:qc_and_t}
\vspace{-2pt}
The generated Q\&A pairs may contain factual errors due to LLM hallucinations~\citep{llm-hulluciation}. To avoid these, we ask a professional multilingual annotation team to verify the quality of the generated Q\&A pairs (the annotation guideline is shown in Figure \ref{guide:qa_verify}). They select examples where the question is natural and answerable and the answer is either correct or correctable by them. For the \textit{monolingual retrieval} setting, we engage a professional translation team to translate the verified Q\&A pairs into German, Spanish, Chinese, and Arabic. For the \textit{multilingual retrieval} setting, translations are performed only into language X for Q\&A pairs derived from X-English article pairs (e.g., into German for Q\&A pair created from German-English article pairs). See Appendix \ref{app:translation} for more details on human translation.

Table \ref{table:stat} presents the dataset statistics after human verification and translation.

\begin{table}[t]
\centering
\Large
\scalebox{0.50}{
\renewcommand{\arraystretch}{1.2}
\begin{tabular}{ll|c|cccc}
\hline
\multicolumn{2}{l|}{\multirow{2}{*}{}}                           & Monolingual Retrieval & \multicolumn{4}{c}{Multilingual Retrieval} \\
\multicolumn{2}{l|}{}                                            & De / Es / Zh / Ar     & De        & Es       & Zh       & Ar       \\ \hline
\multicolumn{2}{l|}{Example Number}                              & 1000                  & 300       & 300      & 300      & 300      \\ \hline
\multicolumn{1}{l|}{\multirow{4}{*}{Question}} & Aggregation     & 313                   & 86        & 99       & 106      & 95       \\
\multicolumn{1}{l|}{}                          & Comparison      & 260                   & 98        & 109      & 81       & 85       \\
\multicolumn{1}{l|}{}                          & Multi-hop       & 215                   & 45        & 40       & 65       & 57       \\
\multicolumn{1}{l|}{}                          & Set             & 212                   & 71        & 52       & 48       & 63       \\ \hline
\multicolumn{1}{l|}{\multirow{2}{*}{Answer}}   & Original        & 872                   & 291       & 296      & 284      & 263      \\
\multicolumn{1}{l|}{}                          & Corrected       & 128                   & 9         & 4        & 16       & 37       \\ \hline
\end{tabular}}
\setlength{\abovecaptionskip}{6pt}
\setlength{\belowcaptionskip}{-12pt}
\caption{XRAG question type statistics for monolingual and multilingual retrieval settings. Answers requiring correction in quality control are also noted.
}
\label{table:stat}
\end{table}

\vspace{-4pt}
\subsection{Selecting the Distracting Articles}
\label{sec:pre_retrieved} 

The grounding articles for each question consist of a set of supporting documents and distracting documents. The two articles used in cross-document question generation serve as supporting documents.  

For distracting documents, we search for documents that are topically related to the question but do not answer it. In the \textit{monolingual retrieval} setting, we use a multilingual dense retriever to search for English documents.   
In the \textit{multilingual retrieval} setting we search for documents in both English and the question language. In both settings we select distracting documents that are published at least two weeks before the supporting articles to ensure that the distracting documents do not answer the question. 

This process yields a set of grounding documents for each question.   In the 
\textit{monolingual retrieval} setting, each question will have six distracting documents and two supporting documents, all in English.
In the \textit{multilingual retrieval} setting, each question will have one supporting document and three distracting documents in English, and the same again in the question language.

\vspace{-2pt}
\section{Benchmarking with XRAG}
\vspace{-1pt}

\vspace{-1pt}
\subsection{Experimental Settings}
\label{sec:exp_setting}
\begin{figure}[t]
\centering\includegraphics[scale=0.41,trim=0 0 0 0]{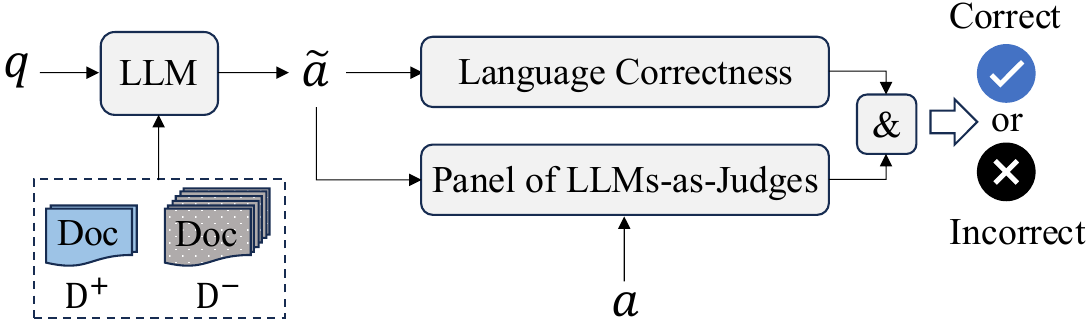}
\setlength{\abovecaptionskip}{-4pt}
\setlength{\belowcaptionskip}{-12pt}
\caption{Evaluation workflow on XRAG: (1) the evaluated LLM generates a response $\tilde{a}$ for a question $q$ based on two supporting articles ${\rm D}^{+}$, and six distracting articles ${\rm D}^{-}$; (2) the response is checked for language correctness; (3) a panel of three LLM judges independently assess the factual accuracy of the response based on the question $q$ and a gold answer $a$, with the final judgment based on majority vote; (4) the final evaluation combines the factual judgment and language correctness.}
\label{fig:evaluation}
\end{figure}

\noindent \textbf{Models}. We benchmark five models on XRAG: GPT-4o~\citep{gpt-4o}, Claude Sonnet-3.5 v1 ~\citep{sonnet3.5}, Mistral-large~\citep{mistral}, Command-R+~\citep{commandrplus}, and Nova Pro~\cite{nova}. These are leading closed- and open-source multilingual LLMs, and have been widely used in RAG research. Unless otherwise specified, the evaluation is conducted by providing the LLM with a question, two supporting documents, and six distracting documents (we show the prompt used to instruct LLMs in using articles to answer questions in Figure \ref{prompt:rag}). Figure \ref{fig:evaluation} shows the evaluation workflow on XRAG.

\noindent \textbf{Evaluation Metrics}. The answers in XRAG are usually simple facts stated in one or two sentences. Following previous work~\citep{crag,domainrag,thakur-etal-2024-knowing}, we use the LLM-as-a-Judge method~\citep{llm-as-judge}, which has proven good at recognizing when two short answers mean the same thing~\citep{kamalloo-etal-2023-evaluating}. To avoid \textit{self-preference}~\citep{panickssery2024llm}, we use a panel of three LLM judges (GPT-4o, Claude Sonnet-3.5, and Mistral-large) with a majority vote. We also use a language detector\footnote{\url{https://github.com/pemistahl/lingua}} to check if the model answer is in the same language as the question; it is marked as incorrect otherwise. To confirm that automatic judging works well, we compare the LLM judge panel decisions with human judges and find a Cohen's kappa score of 0.71. Finally, we report each model's accuracy (\%), as determined by the LLM judge panel, which includes\- the assessment of language correctness. The template used for the LLM-as-a-Judge is shown in Figure \ref{prompt:llmjudge}, and more details regarding the correlation experiments between the LLM judge panel and human evaluations are provided in Appendix~\ref{app:llm-as-a-judge}.

\vspace{-4pt}
\subsection{Establishing QA Performance Bounds}
\label{sec:corpus_ana}
\vspace{-1pt} 
Our goal is to create a cross-lingual RAG benchmark with two key properties: (1) questions should not be answerable using only the parametric knowledge of LLMs, and (2) the task includes challenging questions that require complex reasoning to answer. To assess whether XRAG meets these properties, we evaluate performance of several LLMs on English questions from the \textit{monolingual retrieval} setting of XRAG (see English Q\&A in Figure \ref{fig:construction}) under two conditions: without retrieval, and with the correct supporting articles. Table \ref{table:property} presents the results of models evaluated by the LLM judge panel. All LLMs perform poorly without retrieval, with accuracy rates falling below 16\%. This indicates that these LLMs cannot answer XRAG questions by relying solely on their parametric knowledge. Even when given supporting articles—simulating ideal retrieval— the best result, achieved by GPT-4o,\footnote{N.B.: QA pairs are generated by GPT-4o, and evaluation may be biased in its favor.\label{foot:bias}} reaches only 75.40\% accuracy, which is still far below human accuracy, as we discuss next. These findings show that XRAG questions are challenging even for advanced LLMs.

\begin{table}[t]
\centering
\scalebox{0.8}{
\begin{tabular}{l|cc}
\hline
                  & No Retrieval & Oracle Retrieval \\ \hline
GPT-4o            & \hspace{0.5em}6.30          & 75.40           \\
Claude Sonnet 3.5 & 11.70        & 67.60           \\
Mistral-Large     & 15.20        & 66.40           \\
Command-R+        & 15.30        & 63.50           \\
Nova-Pro          & 13.70        & 68.20           \\ \hline
\end{tabular}}
\setlength{\abovecaptionskip}{6pt}
\setlength{\belowcaptionskip}{-12pt}
\caption{LLM QA accuracy in answering XRAG questions without retrieval, and with XRAG supporting articles  (but without distracting articles).  Questions and supporting articles are in English (see Figure \ref{fig:corpus_analysis}).}
\label{table:property}
\end{table}

\noindent \textbf{Performance Upper Bounds}. To establish a human upper bound on our dataset, we hire human annotators to answer 200 English questions from the \textit{monolingual retrieval} setting (see English Q\&A in Figure \ref{fig:construction}) by carefully reading the article pairs used to create the questions. Their performance is evaluated at 85\% by the LLM judge panel, which is much higher than that of the best LLM.\textsuperscript{\ref{foot:bias}} This shows that even without the cross-lingual challenge, the dataset is a strong benchmark for LLM reasoning. A manual review of answers judged incorrect by the automated evaluator (see Appendix~\ref{app:annotation} for more details on this manual review) finds that 2\% are actually correct, 5\% are wrong with gold answers being correct, and 8\% involve noisy or ambiguous questions. This sets two separate upper bounds: 85\% for human performance, and 92\% allowing for noisy questions.

\begin{table}[t]
\centering
\Large
\scalebox{0.59}{
\renewcommand{\arraystretch}{1.3}
\begin{tabular}{l|c|ccccc}
\toprule 
Doc. Lang.      & \multicolumn{6}{c}{\textbf{En}}          \\
 Query. Lang.    & \textbf{En}    & \textbf{De}    & \textbf{Es}    & \textbf{Zh}    & \textbf{Ar}    & \textbf{Avg.} \\ \midrule
GPT-4o            & 62.40 & 55.90 & 56.80 & 54.70 & 54.70 & 55.50 \\
Claude 3.5 & 42.80 & 37.40 & 40.10 & 37.60 & 38.50 & 38.40 \\
Mistral-large     & 43.30 & 36.50 & 39.50 & 30.60 & 18.90 & 31.40 \\
Command-R+        & 45.70 & 39.80 & 41.20 & 34.30 & 33.80 & 37.30 \\
Nova-Pro          & 54.00 & 44.80 & 49.30 & 37.30 & 34.30 & 41.43 \\ \bottomrule
\end{tabular}
}
\setlength{\abovecaptionskip}{6pt}
\setlength{\belowcaptionskip}{-4pt}
\caption{LLM QA accuracy in the  XRAG {\em monolingual retrieval} setting. Grounding documents consist of two supporting articles and six distracting articles, all in English (see Figure \ref{fig:mono_eval}). QA accuracy with  English queries provides a monolingual RAG baseline for comparison.}
\label{table:mono}
\end{table}

\begin{figure}[t!]
\centering
\includegraphics[scale=0.42,trim=0 0 0 0]{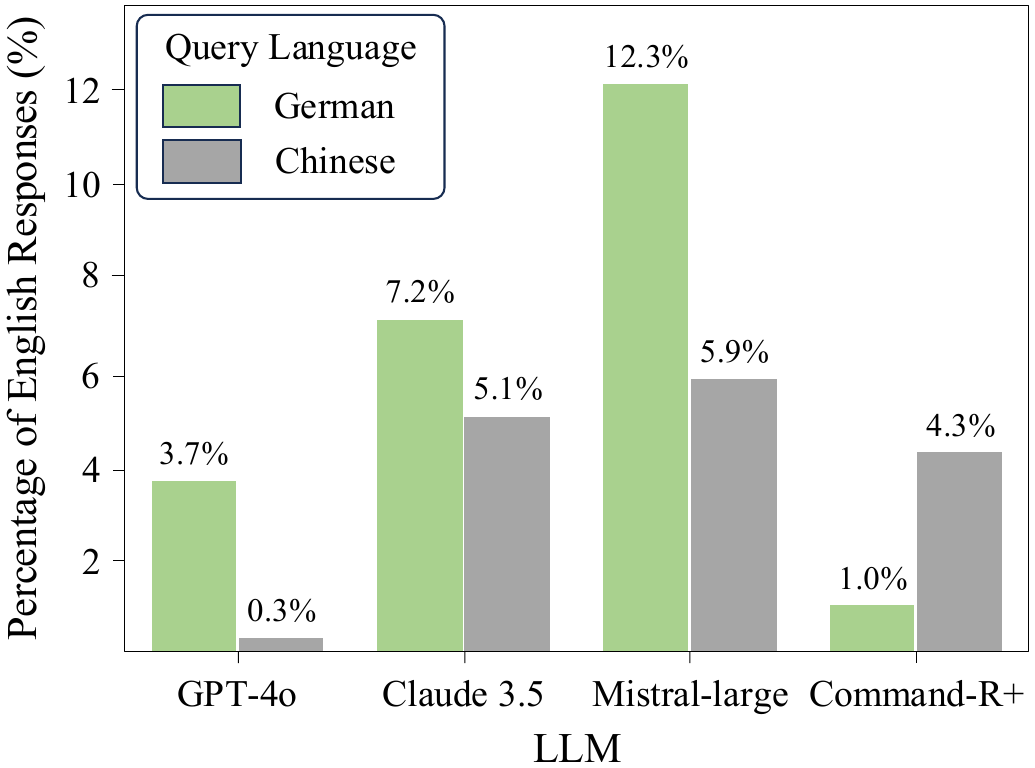}
\setlength{\abovecaptionskip}{4pt}
\setlength{\belowcaptionskip}{-14pt}
\caption{Percentage of instances in cross-lingual RAG  with \textit{monolingual retrieval} (English documents)  where LLMs respond in English instead of the German or Chinese question language.}
\label{fig:rlc}
\end{figure}

\vspace{-4pt}
\subsection{XRAG in Monolingual Retrieval Setting}
\vspace{-1pt}
We first benchmark LLMs in 
cross-lingual RAG with the \textit{monolingual retrieval} setting of XRAG. To highlight the cross-lingual challenges, we compare  results with an English monolingual RAG baseline, where the input question, supporting articles, and distracting articles are all in English. Grounding articles in the \textit{monolingual retrieval} setting are already in English so  this baseline experiment simply   replaces the translated question with its original English  (see English Q\&A in Figure \ref{fig:construction}).

\begin{table*}[!t]
\centering
\scalebox{0.78}{
\begin{tabular}{l|cc|cc|cc|cc|c}
\toprule
Doc. Lang. & \hspace{0.25em}{\bf En+En\textsubscript{\rm De}}\hspace{0.25em} & \hspace{0.25em}{\bf En+De}\hspace{0.25em} & \hspace{0.25em}\textbf{En+En}\textsubscript{\rm Es}\hspace{0.25em} & \hspace{0.25em}{\bf En+Es}\hspace{0.25em} & \hspace{0.25em}{\bf En+En}\textsubscript{\rm Zh}\hspace{0.25em} & \hspace{0.25em}{\bf En+Zh}\hspace{0.25em} & \hspace{0.25em}\textbf{En+En}\textsubscript{\rm Ar}\hspace{0.25em} & \hspace{0.25em}\textbf{En+Ar}\hspace{0.25em} & 
{\bf Avg.}\\
Query. Lang.     & \textbf{En}    & \textbf{De}    & \textbf{En}    & \textbf{Es}    & \textbf{En}    & \textbf{Zh}    & \textbf{En}    & \textbf{Ar}    &  {\bf (crossling.)}                     \\ \midrule
GPT-4o              & 63.33 & 61.67 & 59.33 & 56.00 & 63.00 & 59.33 & 60.67 & 53.33 & 57.58                 \\
Claude 3.5   & 51.00 & 45.67 & 46.33 & 42.67 & 46.67 & 48.00 & 47.33 & 39.67 & 44.00                 \\
Mistral-large       & 45.67 & 42.00 & 43.00 & 39.33 & 48.67 & 37.33 & 43.67 & 32.00 & 37.67                 \\
Command-R+          & 43.67 & 40.00 & 42.33 & 40.33 & 49.67 & 36.33 & 43.33 & 32.00 & 37.17                 \\
Nova-Pro            & 56.33 & 53.00 & 49.67 & 45.33 & 57.67 & 49.33 & 57.33 & 44.67 & 48.08                 \\ \bottomrule
\end{tabular}}
\setlength{\abovecaptionskip}{6pt}
\setlength{\belowcaptionskip}{-10pt}
    \caption{LLM QA accuracy in the XRAG \textit{multilingual retrieval} setting, which for each language {\bf X} consists of a set of questions each accompanied by a supporting document and three distracting documents in language {\bf X} and the same again in English. {\bf En\textsubscript{\rm X}} refers to English translations of documents from language {\bf X} using Google Translate (see Figure \ref{fig:multi_eval}). {\bf En+En\textsubscript{\rm X}} is a monolingual retrieval baseline setting for the language pair {\bf En+X}.}
\label{table:multi}
\end{table*}

Table \ref{table:mono} shows the results as assessed by the LLM judge panel. XRAG in the \textit{monolingual retrieval} setting poses a significant challenge for LLMs, with all models performing poorly. Among them, GPT-4o achieves the highest average accuracy at 55.50\%,\textsuperscript{\ref{foot:bias}} while others score considerably lower, ranging from 31.4\% to 41.43\%. The cross-lingual capabilities of LLMs vary across languages. Compared to the English monolingual RAG baseline, all models experience a performance drop when answering non-English questions, but the severity of this drop differs. GPT-4o and Claude exhibit the smallest and most consistent declines across languages, suggesting more robust multilingual handling. Command-R+ and Mistral show larger variability, indicating potential language-specific weaknesses—particularly for Mistral, which suffers a 56.3\% relative drop in Arabic.

Surprisingly, we find that \textbf{LLMs have issues with Response Language Correctness} (RLC), i.e., they respond in English instead of the question language. Figure \ref{fig:rlc} lists the percentage of cases in the \textit{monolingual retrieval} setting where LLMs respond in the wrong language. GPT-4o and Command-R+ produce the fewest RLC errors, while Mistral-large is most affected.


\vspace{-2pt}
\subsection{XRAG in Multilingual Retrieval Setting}
\label{sec:multi_res}
We now benchmark LLMs in 
cross-lingual RAG with the \textit{multilingual retrieval} setting of XRAG.
As with {\em monolingual retrieval},  we construct an English monolingual RAG setting for comparison. We replace the original non-English questions with their English counterparts (questions before human translation; see English Q\&A in Figure \ref{fig:construction}). We also translate  non-English supporting and distracting articles into English using Google Translate.

Table \ref{table:multi} presents LLM performance in cross-lingual QA in the \textit{multilingual retrieval} setting of XRAG. All models exhibit poor performance in this cross-lingual scenario, with GPT-4o having the highest average accuracy at 57.58\% and Command-R+ the lowest  at 37.17\%. LLMs also show accuracy degradations relative to their corresponding English monolingual RAG baseline, despite the latter being constructed with the assistance of machine translation. 

To identify the most challenging aspect of the cross-lingual RAG with \textit{multilingual retrieval}, we conduct a controlled analysis by gradual conversion to the English monolingual RAG setting. Specifically, we successively replace the question, supporting articles, and distracting articles by their English counterparts from the monolingual RAG baseline. Table \ref{table:control} shows the results of GPT-4o. Changing the question (and expected answer) language from non-English to English only brings a relatively small average  improvement, suggesting that \textbf{non-English generation may not be the core challenge}.\footnote{We also find that LLMs rarely have issues with Response Language Correctness in the \textit{multilingual retrieval} setting.} By contrast, replacing non-English supporting articles with their English translations improves average accuracy noticeably, indicating that \textbf{reasoning over retrieved information across languages is challenging for LLMs}. Translating distracting articles into English also improves performance, implying that identifying useful information in a mixed-language context is harder than from a wholly English one. Similar results are observed with other LLMs, see Appendix \ref{app:control}.

\begin{table}[t]
\centering
\Large
\scalebox{0.59}{
\renewcommand{\arraystretch}{1.25}
\begin{tabular}{l|cccc|c}
\toprule
GPT-4o & \hspace{0.0em}\textbf{En+De}\hspace{0.0em} & \hspace{0.0em}\textbf{En+Es}\hspace{0.0em} & \hspace{0.0em}\textbf{En+Zh}\hspace{0.0em} & \hspace{0.0em}\textbf{En+Ar} & \textbf{Avg.} \\ \midrule
XRAG-MultiR       & 61.67          & 56.00           & 59.33           & 53.33  &   57.58  \\
+EQ           & 63.00          & 52.67           & 60.67           & 56.67    & 58.25      \\
+EQ, +ES       & 65.00          & 57.33           & 62.00           & 60.33 & 61.16 \\
MonoRAG   & 63.33          & 59.33          & 63.00          & 60.67  & 61.58         \\ \bottomrule
\end{tabular}}
\setlength{\abovecaptionskip}{7pt}
\setlength{\belowcaptionskip}{-12pt}
\caption{Controlled analysis of GPT-4o on the \textit{multilingual retrieval} setting of XRAG, replacing questions (EQ), supporting articles (ES), and distracting articles (ED) with their English counterparts from the English monolingual RAG settings (see Figure \ref{fig:controlled_analysis}). "XRAG-MultiR" is the \textit{multilingual retrieval} setting, and "MonoRAG" (+EQ, +ES, +ED) is the English monolingual RAG baseline setting.}
\label{table:control}
\end{table}

\vspace{-4pt}
\section{Conclusions}
\vspace{-2pt}
We introduce XRAG, a benchmark for evaluating the generation abilities of LLMs in cross-lingual RAG settings. We introduce a novel LLM-based workflow for creating questions that require complex reasoning and external documents to answer. Experiments reveal that LLMs significantly underperform humans on XRAG—even without cross-lingual elements—highlighting its utility for assessing reasoning ability. Further analysis shows that LLMs struggle with response language correctness in the XRAG \textit{monolingual retrieval} setting and with reasoning over retrieved content across languages in the XRAG \textit{multilingual retrieval} setting.

\section{Limitations}
Our work has some limitations. First, our multilingual retrieval setting solely covers the scenario of two languages (English and the question language). However, there may be cases involving retrieval across a set of languages (more than two). Note that our construction pipeline can support exploration in this setup by using more articles to generate questions, which we leave for future work. Second, we only benchmarked five models in this work because of legal concerns. It would be interesting to see how other LLMs perform on XRAG. Finally, we could conduct more insightful controlled analyses on XRAG, such as exploring the impact of the number of distracting articles. Due to space limitations, we leave this for future work.


\bibliography{main.bbl}

\clearpage
\appendix
\begin{table}[t]
    \centering
    \scalebox{0.85}{
    \begin{tabular}{l|c|c|c}
        \hline
        User Locale & Native & English & Both \\
        \hline
        Germany & 59.8\% & 56.3\% & 71.5\%\\
        Japan & 61.1\% & 44.3\% & 68.3\% \\
                Spain & 57.9\% & 48.4\% & 68.9\%\\

        \hline
    \end{tabular}}
    \setlength{\abovecaptionskip}{8pt}
    \setlength{\belowcaptionskip}{-14pt}
    \caption{Percentage of satisfactory retrieval results using Google search as retriever on real-world information seeking LLM traffic. Search results are evaluated by Claude 3.5 Sonnet if they contain sufficient information for a satisfactory answer to the user query. \emph{Native} uses Google search in the user locale, \emph{English} performs English Google search using a translation of the user query, and \emph{Both} combines the two search results. LLM traffic is obtained via SimilarWeb.}
    \label{table:sony}
\end{table}

\section{Retrieval Quality Investigation}
\label{app:sony}

We evaluate the relevance of English and native-language search results for real-world queries submitted to large language models (LLMs) by non-English users in Germany, Spain, and Japan. The analysis is based on a proprietary dataset of real LLM traffic provided by SimilarWeb\footnote{https://www.similarweb.com/}. User queries are filtered by locale and language using langid to retain only those submitted in the respective native languages. For each query, we retrieve two sets of search results: one from a U.S.-based Google search using the English translation of the query, and another from a country-specific Google domain using the original native-language version. Relevance is assessed using Claude 3.5 Sonnet, which evaluates whether the retrieved results contain sufficient information to generate a satisfactory response, taking into account the user’s locale (e.g., a tax-related query from Germany must include references to German tax regulations). Table \ref{table:sony} reports the percentage of queries for which the English-only, native-only, or both sets of results independently provided sufficient information. The results indicate that combining English and native-language search results significantly improves the proportion of queries for which a comprehensive response can be generated, compared to using either language alone.

\section{Data Construction}
\label{app:data_construct}

\subsection{Data Source}
\label{app:data_source}
We use articles in News Crawl~\citep{newscrawl} as the data source to create questions and prepare supporting and distracting articles. Specifically, we download news articles between June 1, 2024 and November 30, 2024 from NEWS Crawl using new-please\footnote{https://github.com/fhamborg/news-please}. This timeframe exceeds the knowledge cutoff of widely used LLMs, such as GPT-4o and Claude 3.5 sonnet. Therefore, questions created from these articles are more likely to require LLMs to use external knowledge to answer. We only keep articles that contain more than 1200 tokens and are in English, German, Spanish, Chinese, or Arabic, obtaining around 1700k, 250k, 600k, 180k, and 460k news articles in these languages.

\begin{figure}[t]
\centering\includegraphics[scale=0.48,trim=0 0 0 0]{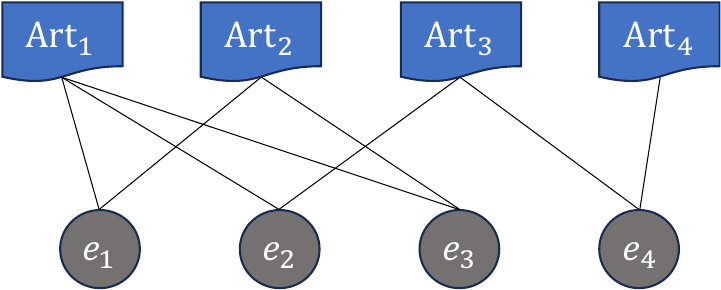}
\setlength{\abovecaptionskip}{8pt}
\setlength{\belowcaptionskip}{-12pt}
\caption{Example of a bipartite graph between articles and entities.}
\label{fig:bipartite}
\end{figure}

\subsection{Identify Pairs of Related Articles}
\label{app:find_pairs}
\subsubsection{Identify English-English Article pairs}
\label{app:find_en_en}
Inspired by the concept of "bridge entity" in~\citet{yang-etal-2018-hotpotqa}, we use a bipartite graph between articles and entities to find related English article pairs. Specifically, we randomly sample around 100k English articles from the data downloaded from News Crawl, covering various topics such as Politics, Sports, Economy, and Entertainment. Then, we use stanza~\citep{qi-etal-2020-stanza} to identify entities in the titles of the sampled articles and construct a bipartite graph between the articles and entities (as shown in Figure \ref{fig:bipartite}). In this graph, nodes are entities and articles, and an edge will be added between an entity and an article if the entity is contained in the title of the article. Finally, we perform the Depth-First Search on the graph to find pairs of articles sharing at least two entities in their titles (e.g., articles 1 and 2 in Figure \ref{fig:bipartite}) and having a publication time gap of no more than two weeks. Here we use the title instead of the main text because entities in the title are often the key entities of the news, and two articles containing the same key entities are more likely to be related. 

\subsubsection{Identify X-English Article Pairs}
\label{app:find_en_no_en}
To construct cross-document questions for the \textit{multilingual retrieval} setting in XRAG, we need to find related articles across languages. We empirically find that the article-entity graph performs poorly here, due to (1) inaccurate entity recognition in non-English texts~\citep{malmasi-etal-2022-semeval}, and (2) challenges in cross-lingual entity mapping~\citep{liang-etal-2024-addressing}.

\begin{figure}[t]
\centering
\includegraphics[scale=0.43,trim=0 0 0 0]{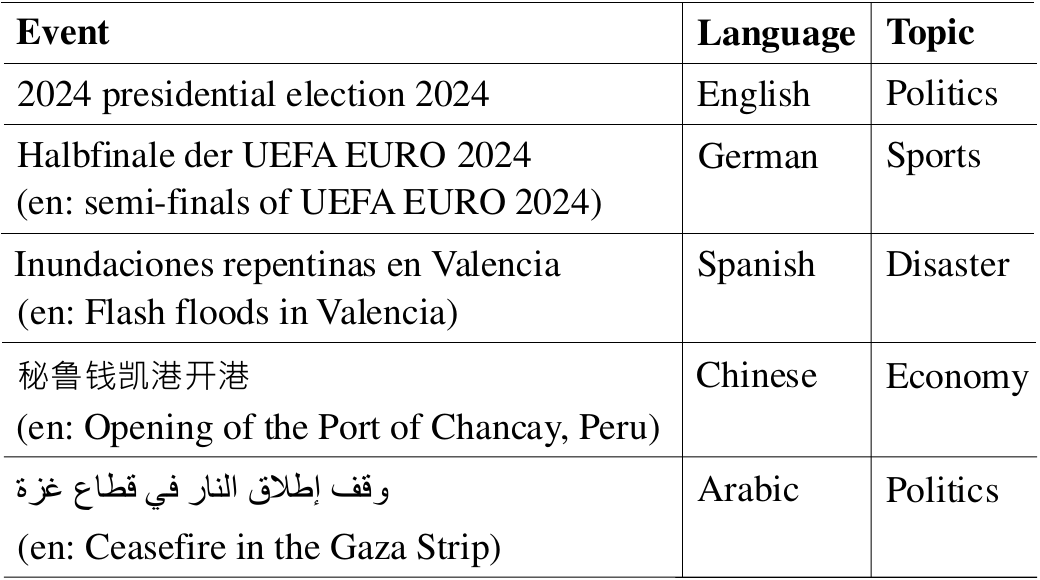}
\setlength{\abovecaptionskip}{-4pt}
\setlength{\belowcaptionskip}{0pt}
\caption{Examples of events collected from Wiki 2024 between June and November, which will be used to retrieve related articles across languages.}
\label{fig:events}
\end{figure}

\begin{figure}[t]
\centering\includegraphics[scale=0.41,trim=0 0 0 0]{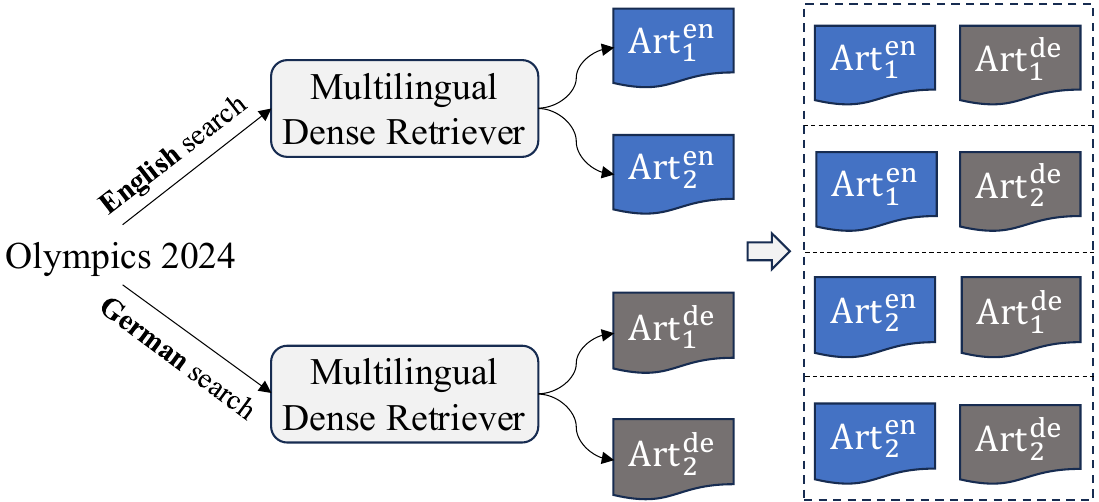}
\setlength{\abovecaptionskip}{-4pt}
\setlength{\belowcaptionskip}{-12pt}
\caption{Example of using the event "Olympics 2024" and a multilingual retrieve to find related articles in English and German.}
\label{fig:retrieve_related}
\end{figure}

To address this, we collect 117 international events between June and November 2024 from different language versions of Wiki 2024\footnote{
\url{https://en.wikipedia.org/wiki/2024} \\
\url{https://de.wikipedia.org/wiki/2024} \\
\url{https://es.wikipedia.org/wiki/2024} \\
\url{https://zh.wikipedia.org/zh-cn/2024} \\
\url{https://ar.wikipedia.org/wiki/2024}}, covering topics such as politics, sports, astronomy, and natural disasters (see some examples in Figure \ref{fig:events}). We also build a multilingual dense retriever\footnote{The dense retriever is also used in selecting distracting articles for each question, see Section \ref{sec:pre_retrieved}.} based on the multilingual model BGE-M3~\citep[as text encoder,][]{chen-etal-2024-m3} and Milvus~\citep[as vector database,][]{milvus}. The retriever operates on news articles downloaded from NEWS Crawl. Then, we use the retriever to search across languages for articles related to the events. Finally, we group articles in different languages about the same event to form related article pairs.  Figure \ref{fig:retrieve_related} show an example, where we use the event "Olympics 2024" to search for English and German articles and create related English-German article pairs.

\begin{figure}[t]
\centering\includegraphics[scale=0.37,trim=0 0 0 0]{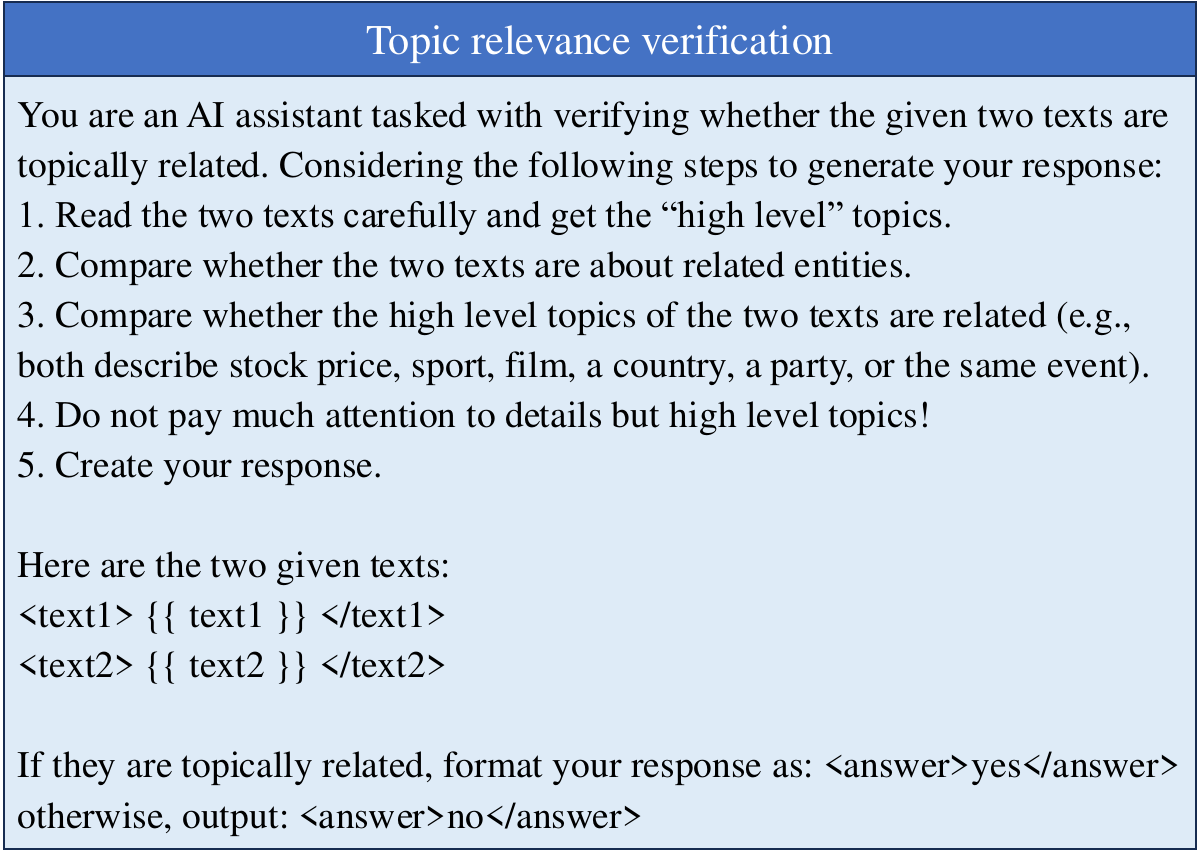}
\setlength{\abovecaptionskip}{6pt}
\setlength{\belowcaptionskip}{-2pt}
\caption{Prompt for topic relevance verification.}
\label{prompt:topic_verify}
\end{figure}

\begin{figure}[t]
\centering
\includegraphics[scale=0.37,trim=0 0 0 0]{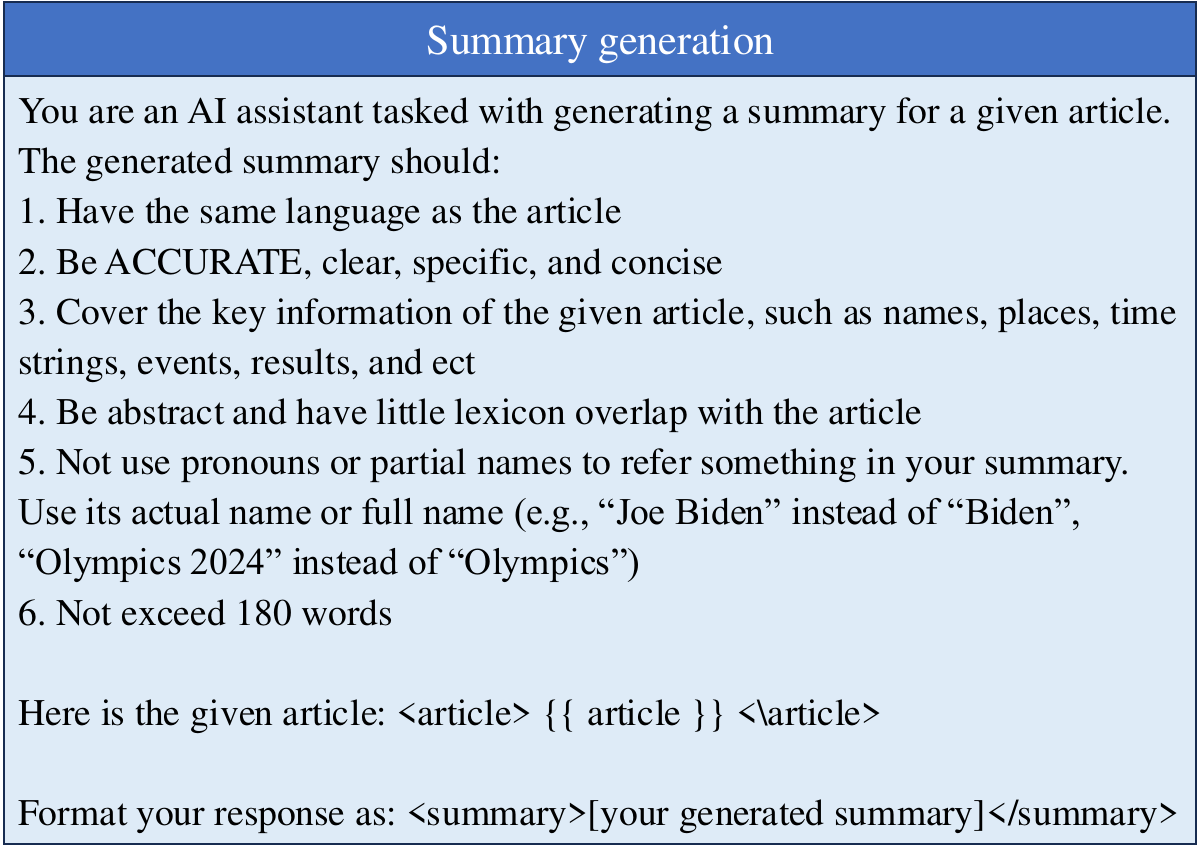}
\setlength{\abovecaptionskip}{6pt}
\setlength{\belowcaptionskip}{-12pt}
\caption{Prompt for summary generation, as described in Step 1 of Section \ref{sec:q_generation}.}
\label{prompt:summary}
\end{figure}

\begin{figure}[!t]
\centering
\includegraphics[scale=0.37,trim=0 0 0 0]{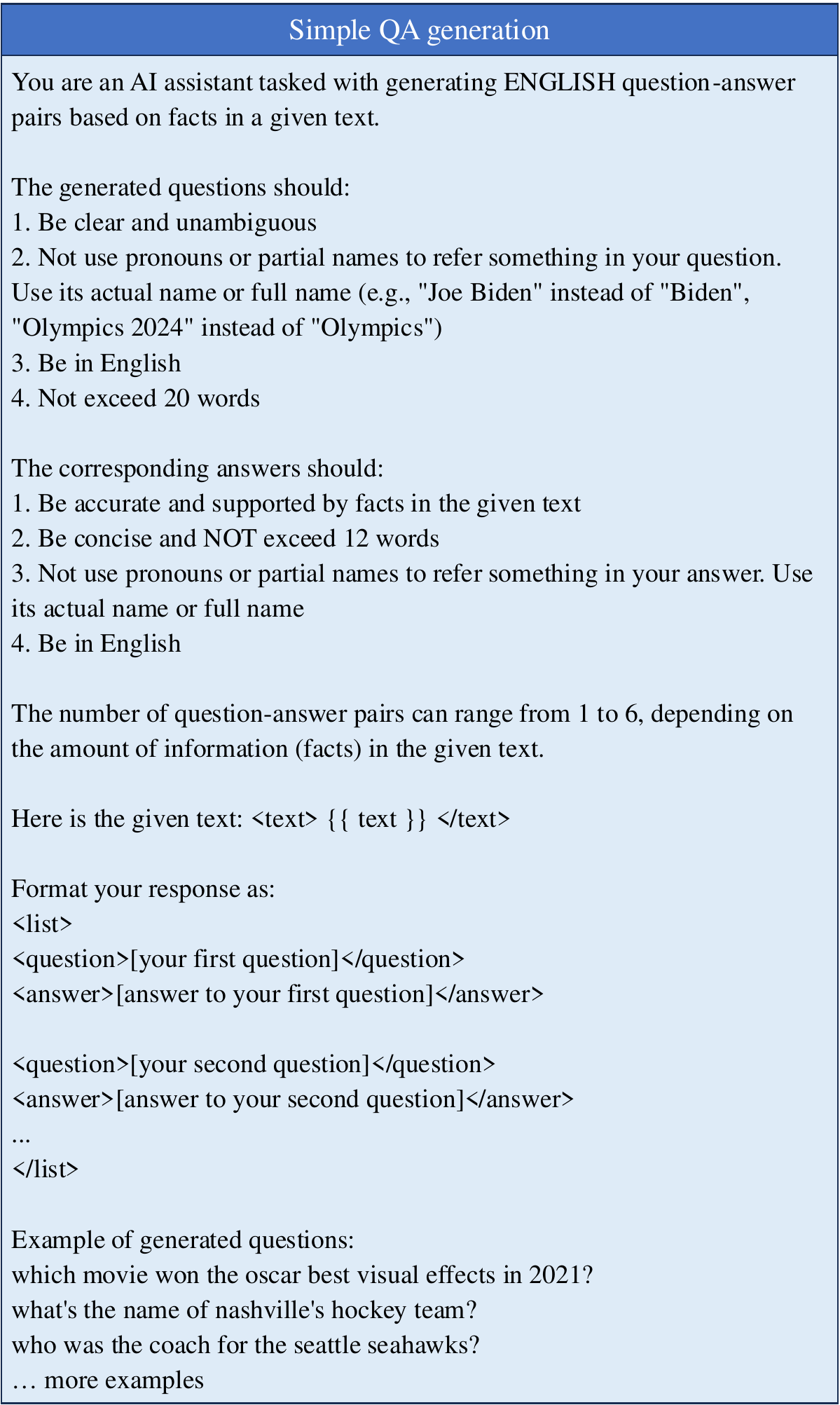}
\setlength{\abovecaptionskip}{6pt}
\setlength{\belowcaptionskip}{-16pt}
\caption{Prompt for simple Q\&A generation, as described in Step 2 of Section \ref{sec:q_generation}.}
\label{prompt:simple_qa}
\end{figure}

\begin{figure}[!t]
\centering
\includegraphics[scale=0.37,trim=0 0 0 0]{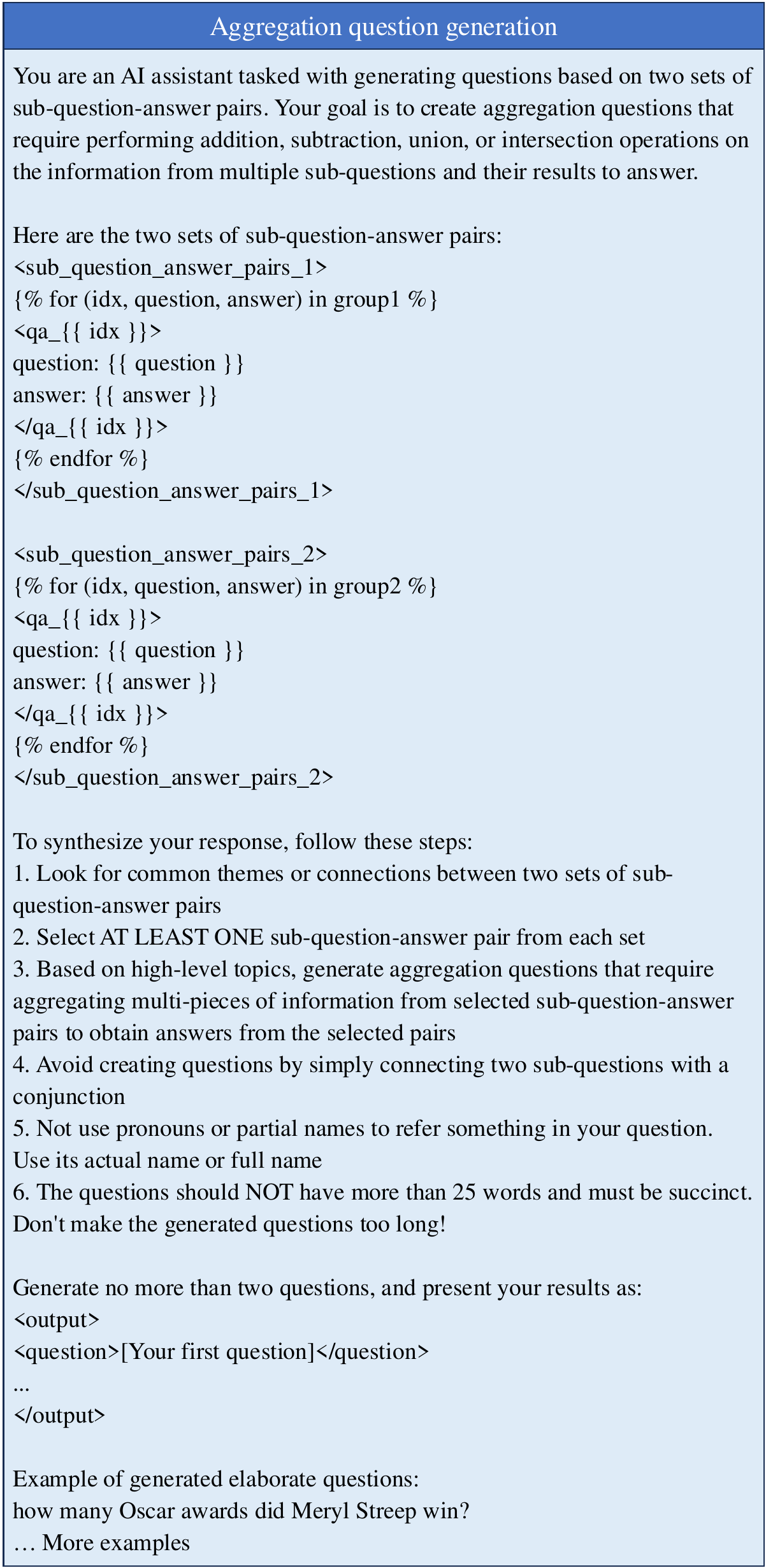}
\setlength{\abovecaptionskip}{-6pt}
\setlength{\belowcaptionskip}{-16pt}
\caption{Prompt for aggregation question generation, as described in Step 3 of Section \ref{sec:q_generation}.}
\label{prompt:aggregation}
\end{figure}

\begin{table*}[t]
\centering
\Large
\renewcommand{\arraystretch}{1.3}
\scalebox{0.56}{
\begin{tabular}{l|l}
\hline
\textbf{Quesiton type} & \textbf{Definition}                                                                                                                                                           \\ \hline
Aggregation            & \begin{tabular}[c]{@{}l@{}}Questions that require the aggregation of information across articles to answer (e.g., “how many Oscar awards \\ did Meryl Streep win?”)\end{tabular}             \\ \hline
Comparison             & \begin{tabular}[c]{@{}l@{}}Questions that compare information in two articles (e.g., “who started performing earlier, Adele or Ed Sheeran?”)\end{tabular}                                  \\ \hline
Multi-hop              & \begin{tabular}[c]{@{}l@{}}Questions that require chaining multiple pieces of information from two articles to compose the answer (e.g., \\ “who acted in Ang Lee’s latest movie?”)\end{tabular} \\ \hline
Set                    & \begin{tabular}[c]{@{}l@{}}Questions that expect a set of entities, objects, or events from two articles as the answer (e.g., “what are the continents \\ in the southern hemisphere?”)\end{tabular}      \\ \hline
\end{tabular}}
\setlength{\abovecaptionskip}{4pt}
\setlength{\belowcaptionskip}{-10pt}
\caption{Definition of four types of cross-document questions: aggregation, comparison, multi-hop, and set questions, as mentioned in Step 3 of Section \ref{sec:q_generation}.}
\label{table:q_types}
\end{table*}

In both cases, we further use GPT-4o to verify the topical relevance of the found article pairs, passing only those confirmed to be truly related to the next step. Figure \ref{prompt:topic_verify} shows the prompt we used to instruct GPT-4o for relevancy verification. Furthermore, if computational resources were not a constraint, a more generalized approach to identifying pairs of related articles would involve using randomly selected documents as queries and retrieving related pairs through a multilingual dense retriever. This method would be applicable across a wide range of domains and languages.


\begin{figure}[t]
\centering
\includegraphics[scale=0.36,trim=0 0 0 0]{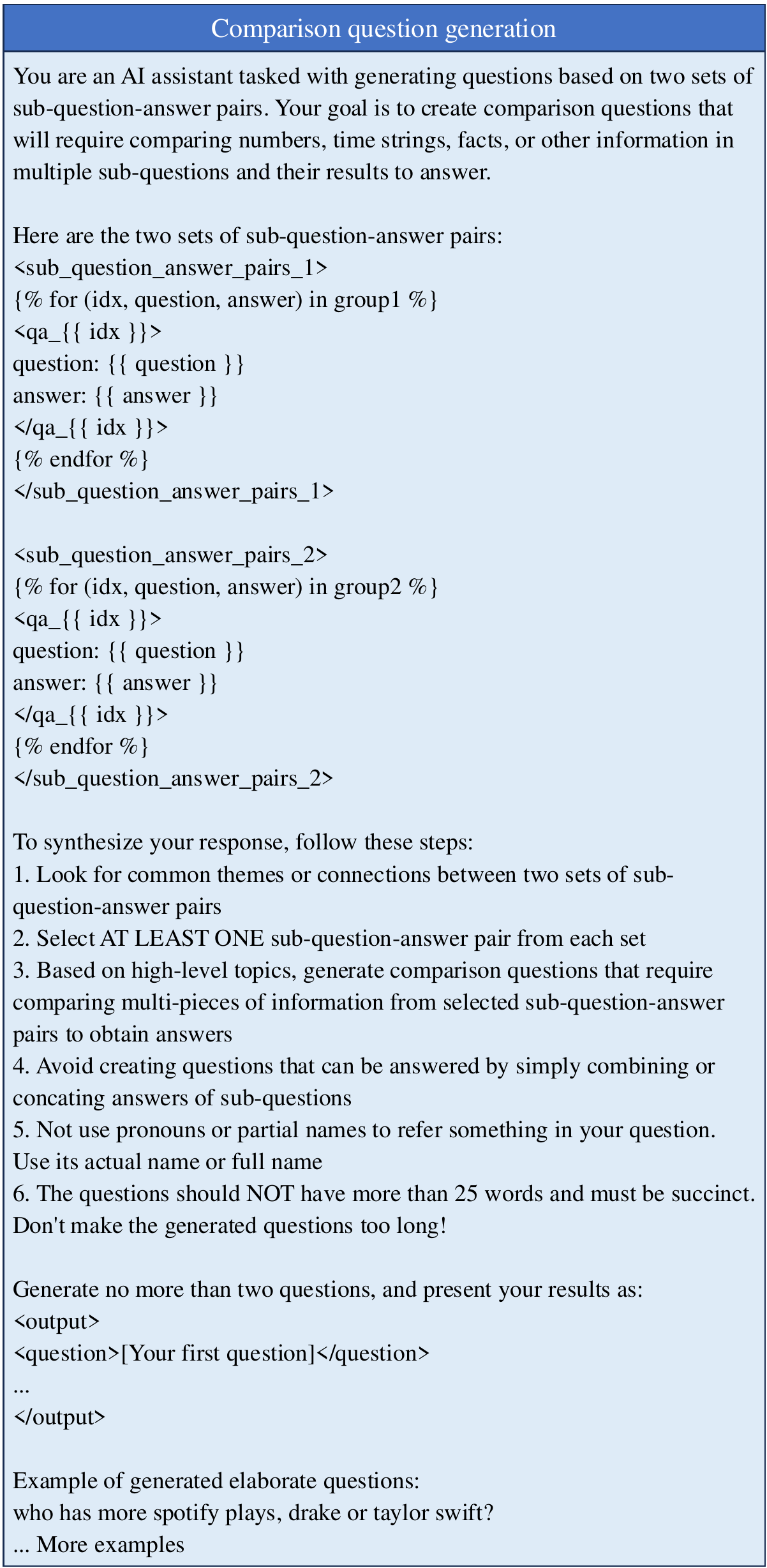}
\setlength{\abovecaptionskip}{5pt}
\setlength{\belowcaptionskip}{-15pt}
\caption{Prompt for comparison question generation, as described in Step 3 of Section \ref{sec:q_generation}.}
\label{prompt:comparison}
\end{figure}

\subsection{English cross-document Q\&A Generation}
\label{app:en_qa}

\begin{figure}[t]
\centering
\includegraphics[scale=0.36,trim=0 0 0 0]{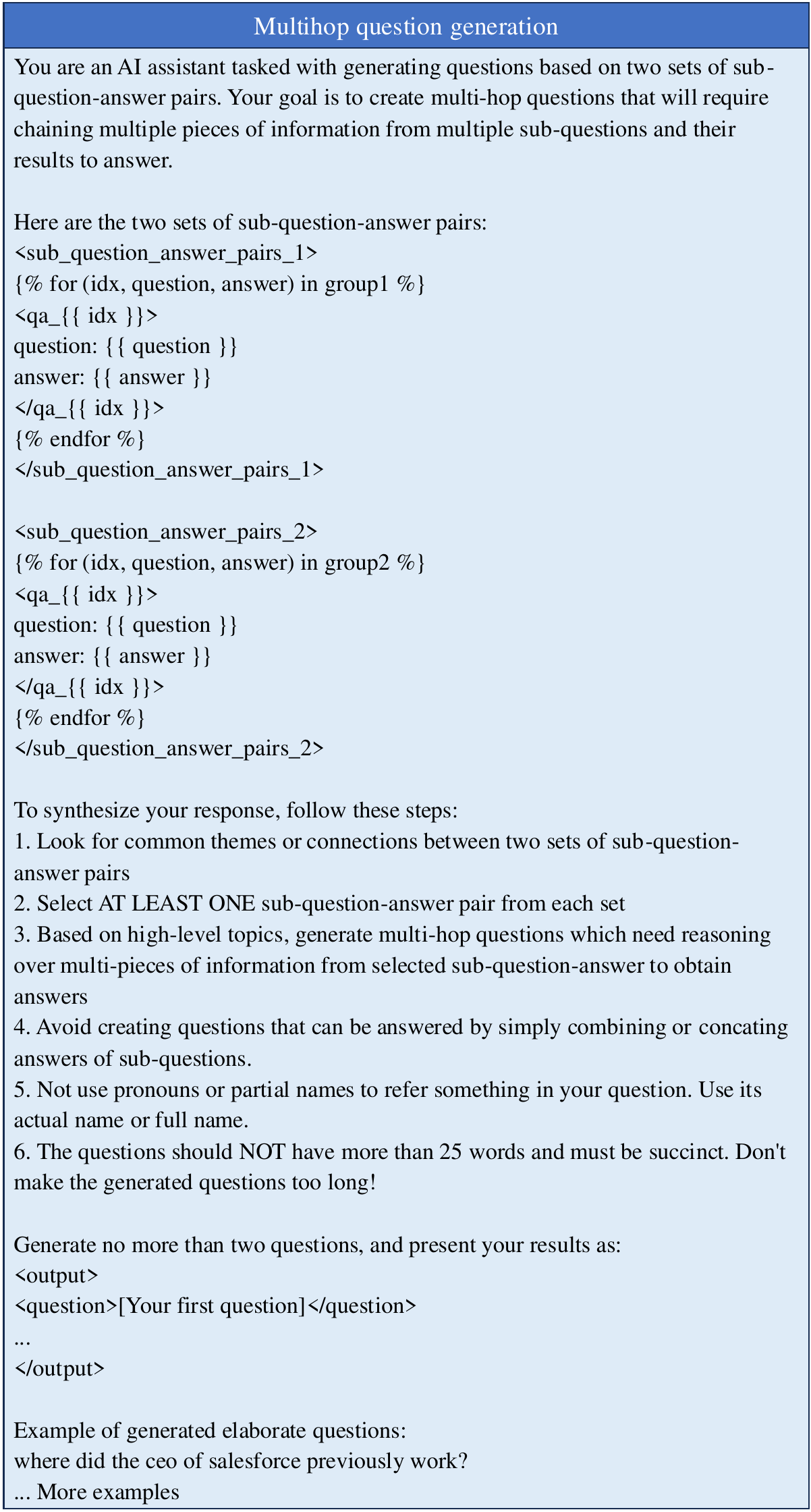}
\setlength{\abovecaptionskip}{6pt}
\setlength{\belowcaptionskip}{-12pt}
\caption{Prompt for multi-hop question generation, as described in Step 3 of Section \ref{sec:q_generation}.}
\label{prompt:multihop}
\end{figure}

\begin{figure}[t]
\centering
\includegraphics[scale=0.37,trim=0 0 0 0]{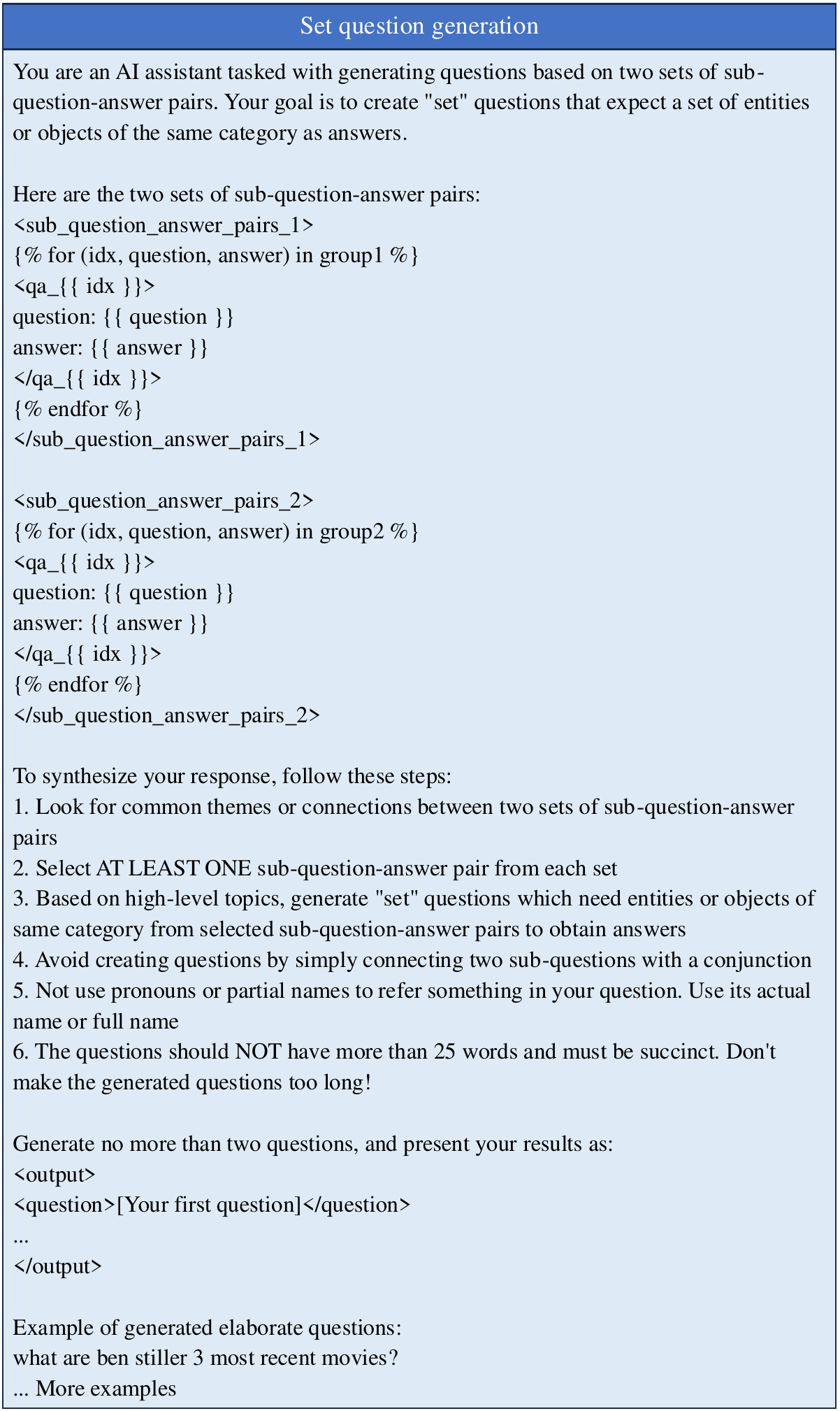}
\setlength{\abovecaptionskip}{0pt}
\setlength{\belowcaptionskip}{-16pt}
\caption{Prompt for set question generation, as described in Step 3 of Section \ref{sec:q_generation}.}
\label{prompt:set}
\end{figure}

\begin{figure}[t]
\centering\includegraphics[scale=0.37,trim=0 0 0 0]{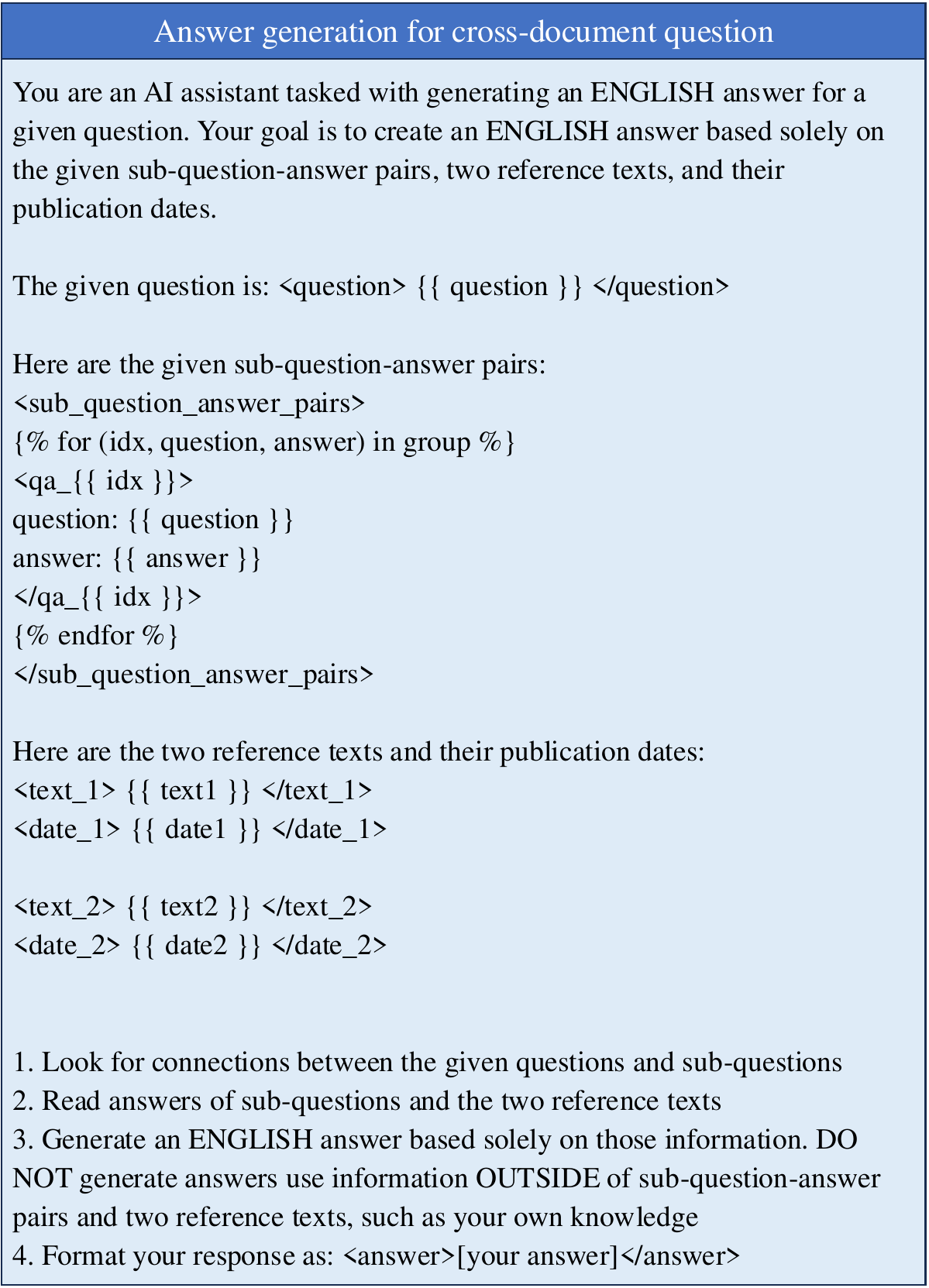}
\setlength{\abovecaptionskip}{8pt}
\setlength{\belowcaptionskip}{-16pt}
\caption{Prompt for creating an answer for a generated cross-document question, as described in Step 3 of Section \ref{sec:q_generation}.}
\label{prompt:crossdoc_a}
\end{figure}

Figure \ref{prompt:summary} shows the prompt used to generate a summary from an article, as described in Step 1 of Section \ref{sec:q_generation}. Figure \ref{prompt:simple_qa} shows the prompt for generating a set of simple Q\&A pairs from a summary, as described in Step 2 of Section \ref{sec:q_generation}. Table \ref{table:q_types} shows the definition of four types of cross-document questions: aggregation, comparison, multi-hop, and set questions, as mentioned in Step 3 of Section \ref{sec:q_generation}. We show the prompts used for generating these four types of questions in Figures \ref{prompt:aggregation}, \ref{prompt:comparison}, \ref{prompt:multihop}, and \ref{prompt:set}. The prompt used for generating an answers to the created cross-document question is presented in Figure \ref{prompt:crossdoc_a}.

\begin{figure*}[t]
\centering
\includegraphics[scale=0.44,trim=0 0 0 0]{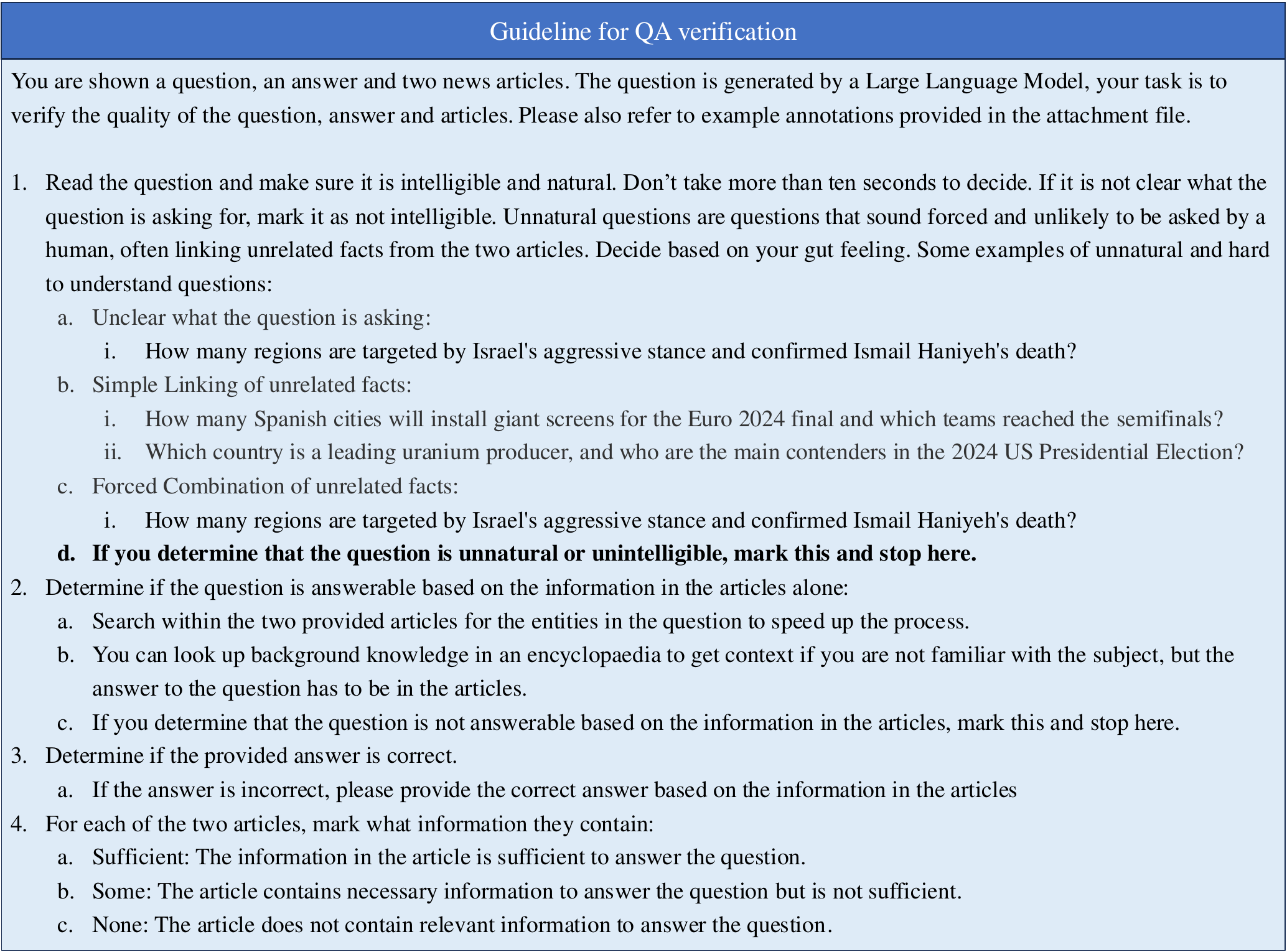}
\setlength{\abovecaptionskip}{6pt}
\setlength{\belowcaptionskip}{-8pt}
\caption{Guidelines for verifying the quality of generated cross-document Q\&A pairs, as described in Section \ref{sec:qc_and_t}. We also provide additional examples to guide the annotation.}
\label{guide:qa_verify}
\end{figure*}

\begin{figure}[!t]
\centering
\includegraphics[scale=0.37,trim=0 0 0 0]{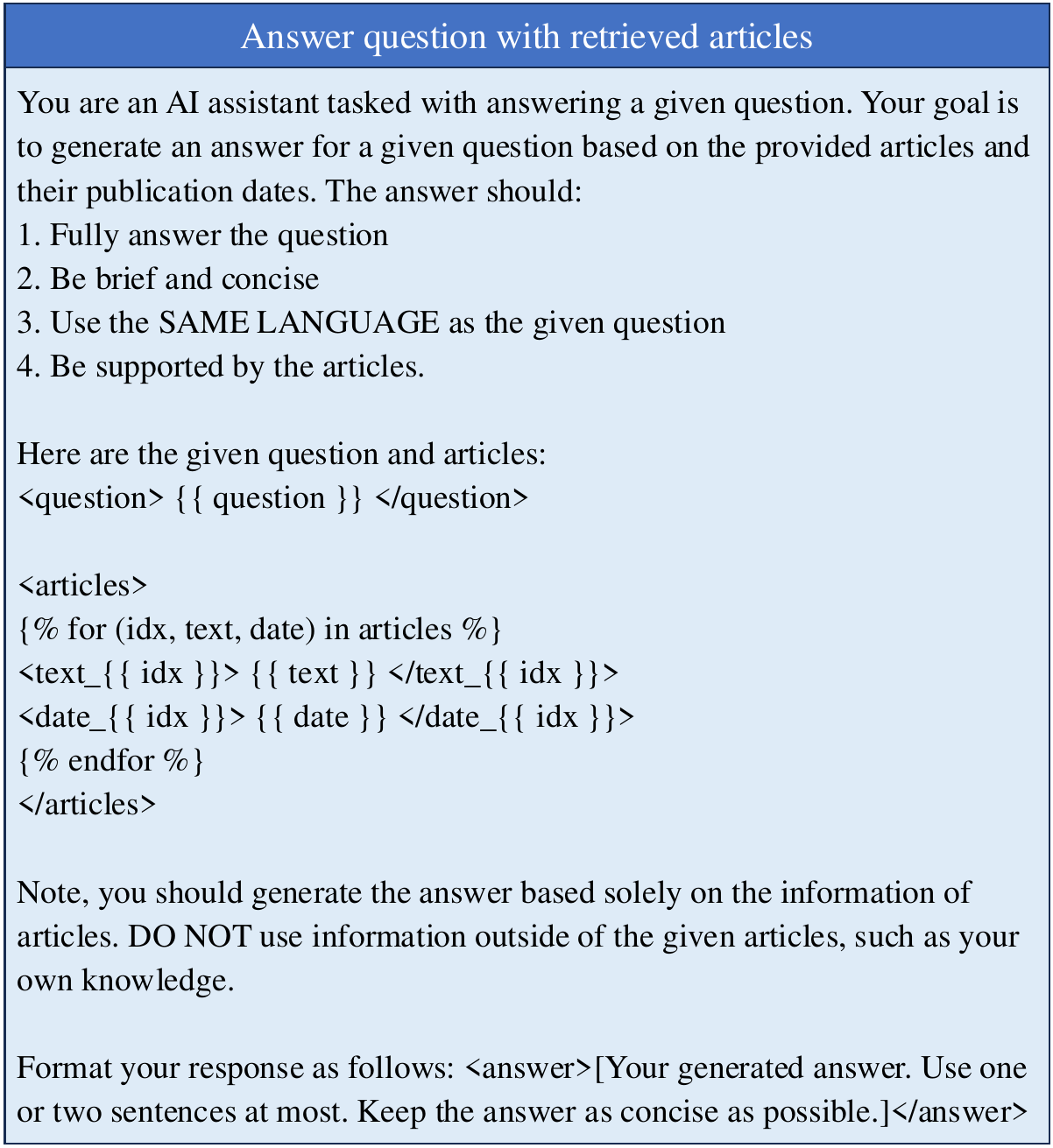}
\setlength{\abovecaptionskip}{0pt}
\setlength{\belowcaptionskip}{-16pt}
\caption{Prompt used to instruct LLMs in using articles to answer questions, as described in Section \ref{sec:exp_setting}.}
\label{prompt:rag}
\end{figure}

\begin{figure}[!t]
\centering
\includegraphics[scale=0.37,trim=0 0 0 0]{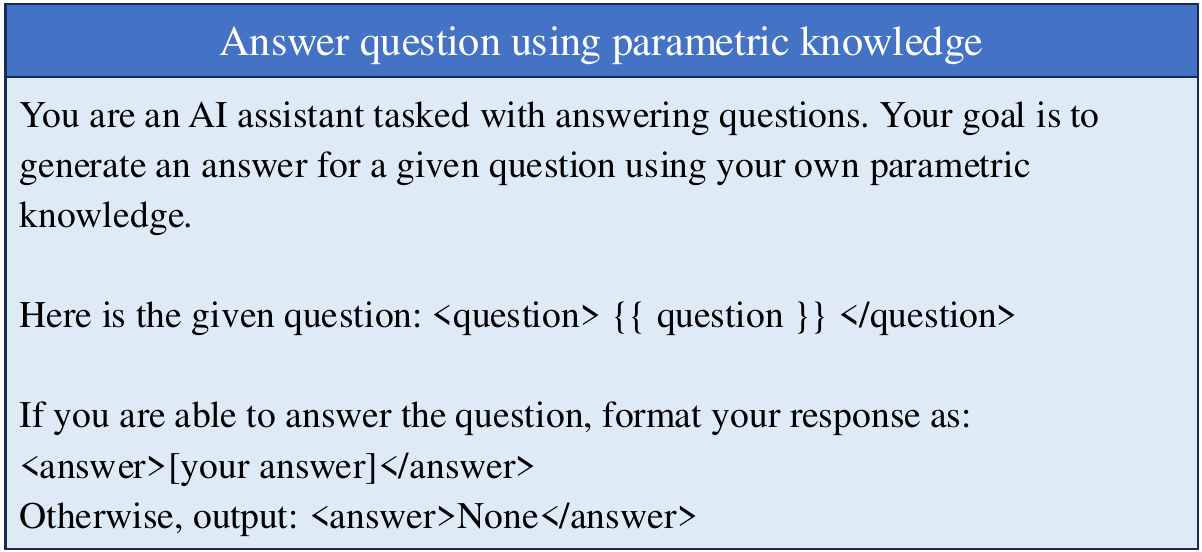}
\setlength{\abovecaptionskip}{0pt}
\setlength{\belowcaptionskip}{-16pt}
\caption{Prompt for answering question using parametric knowledge of LLMs. This corresponds to the "No Retrieval" setting in Section \ref{sec:corpus_ana} and Table \ref{table:property}.}
\label{prompt:no_rag}
\end{figure}

\subsection{Human Verification}
\label{app:annotation}
Due to the existence of LLM hallucinations~\citep{llm-hulluciation}, the Q\&A pairs generated by the LLM may contain factual errors. Therefore, we ask a professional annotation team to verify the quality of the generated Q\&A pairs. Figure \ref{guide:qa_verify} presents the guidelines we prepared for the annotation team. 

We generate 2,950 raw cross-document Q\&A pairs from English article pairs. Following manual verification, approximately 90\% of the questions are deemed natural, 72\% are considered answerable, and 54\% of the answers (including some that were manually corrected) are judged correct, yielding a final set of approximately 1,500 verified Q\&A pairs. To further assess quality, we sample 200 Q\&A pairs and task a separate group of annotators to answer the questions by carefully reading the article pairs used to generate these questions. Annotators are explicitly instructed to answer no more than one question per hour (to ensure thorough reading and accurate responses). If their answers align with the reference answers, it suggests high-quality Q\&A pairs. We use the majority vote of three LLM-as-a-Judge (see Evaluation Metrics in Section \ref{sec:exp_setting}) to calculate the accuracy of human responses to 200 questions, resulting in 85\%. Upon manual review of the 30 failed cases, we find: (i) 4 human answers are correct but incorrectly judged by the LLMs; (ii) 10 are genuinely incorrect, with the reference answers being valid; and (iii) 16 are difficult to assess due to ambiguity or poor question quality, and are thus categorized as low-quality examples. This performance is comparable to established QA benchmarks—for instance, human accuracy on SQuAD~\citep{rajpurkar-etal-2016-squad} is 86.8\%. We then randomly sample 1000 examples from the 1500 Q\&A pairs and send them for human translation.

We generate approximately 1,000 Q\&A pairs each from article pairs of the following language pairs: English–German, English–Spanish, English–Chinese, and English–Arabic. After manual verification, about 90\% of the questions from the English–German and English–Spanish sets are deemed natural, while the proportion for English–Chinese and English–Arabic is slightly lower, at approximately 84\%. Finally, we obtain 487, 680, 332, and 420 high-quality Q\&A pairs from the English–German, English–Spanish, English–Chinese, and English–Arabic article pairs, respectively. From each set, 300 Q\&A pairs are randomly sampled and submitted for translation.

\subsection{Human Translation}
\label{app:translation}
Q\&A pairs generated from English-English article pairs are translated into German, Spanish, Chinese, and Arabic to simulate a cross-lingual retrieval-augmented generation (RAG) with \textit{monolingual retrieval}. Given the importance of named entities in Q\&A, translators are instructed to consult Wikipedia or other reliable sources to find commonly used translations in the target language. If no appropriate translation exists, the original English term is retained. For example, "Microsoft updated the Copilot" is translated into Chinese as "\zh{微软更新了Copilot}," where "Microsoft" is translated (\zh{微软}) and "Copilot" remains in English due to the absence of a standard Chinese equivalent. 

For Q\&A pairs generated from article pairs in different languages, such as English-German and English-Chinese, translation is performed only into the language of the non-English input article. For example, Q\&A pairs from English–German article pairs are translated into German, and so on for others. These examples are used to simulate a cross-lingual RAG with \textit{multilingual retrieval}.

\vspace{-5pt}
\section{Experimental Settings}
\vspace{-5pt}
\subsection{Models}
\label{app:models}
\vspace{-2pt}
We benchmark five models on XRAG, including

\noindent GPT-4o-2024-08-06, Claude 3.5 Sonnet (2024-06-20), Mistral-Large-Instruct-2407, Command-r+, and Nova-pro. Figure \ref{prompt:rag} shows the template we use to prompt LLMs to respond to a given answer by reading the retrieved articles.  Figure \ref{prompt:no_rag} shows the template used to prompt LLMs to answer questions using their own parametric knowledge.

\subsection{Evaluation Metrics}
\label{app:llm-as-a-judge}
\begin{figure}[t]
\centering
\includegraphics[scale=0.36,trim=0 0 0 0]{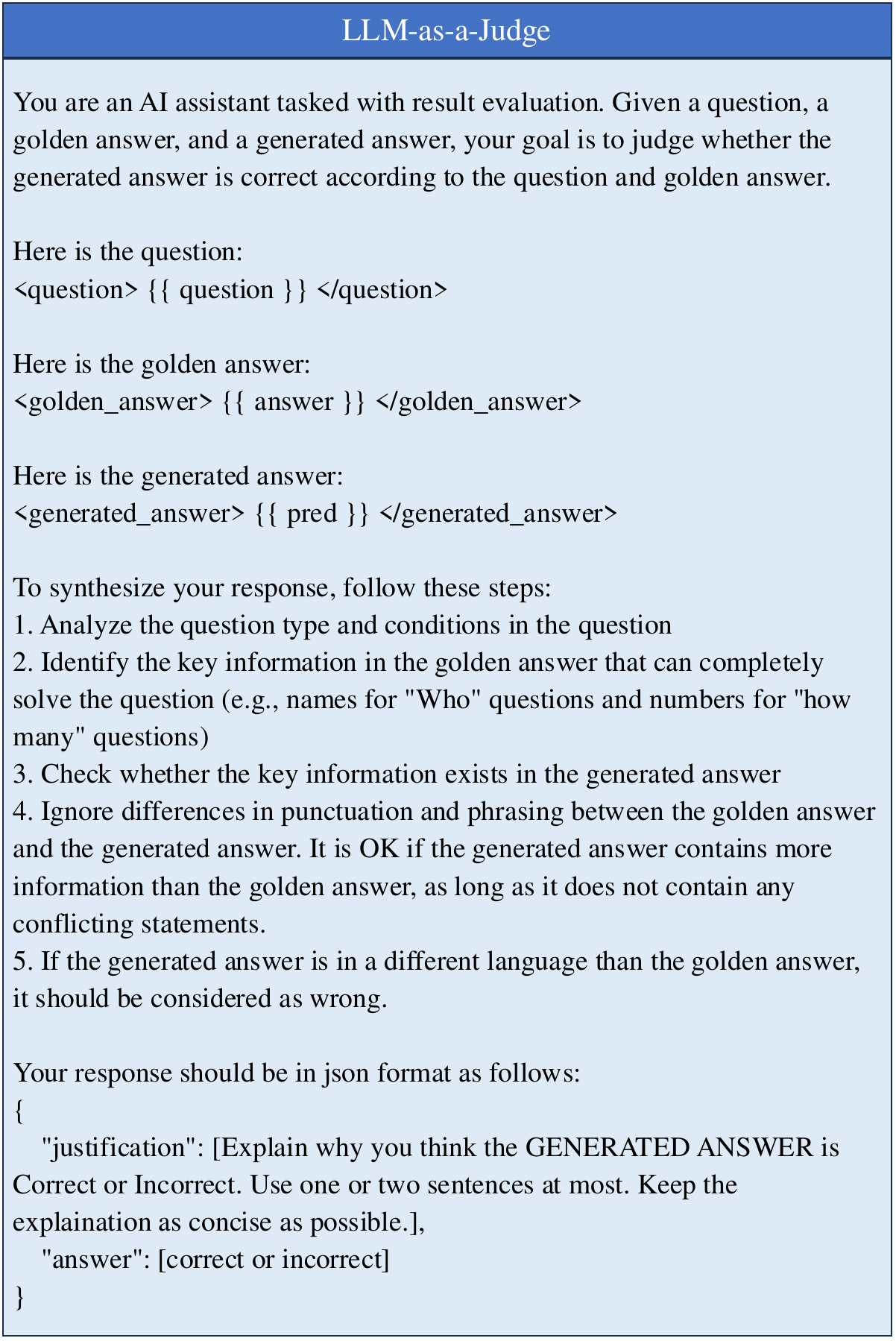}
\setlength{\abovecaptionskip}{6pt}
\setlength{\belowcaptionskip}{0pt}
\caption{Prompt used for LLM-as-a-Judge, as described in Section \ref{sec:exp_setting}.}
\label{prompt:llmjudge}
\end{figure}

We use LLM-as-a-Judge to determine whether an LLM's response is correct. Specifically, each time we input a question, a golden answer, and an answer generated by a model to an LLM and ask the LLM to determine whether the generated answer is correct or incorrect following the guideline we provided (see prompt in Figure \ref{prompt:llmjudge}). To avoid the \textit{self-preference} problem~\citep{panickssery2024llm}, we use three LLM judges, including GPT-4o-2024-08-06, Claude Sonnet-3.5 (2024-06-20), and Mistral-Large-Instruct-2407, and take the majority vote as the final result. In the prompt, we explicitly instruct LLM judges to consider the language of models' responses, but they sometimes fail to do so. To address this, we apply a language detection tool, lingua\footnote{\url{https://github.com/pemistahl/lingua}}, to verify whether the response is in the same language as the corresponding input question, and if not, we consider it incorrect. Finally, we report each model's accuracy by the LLM judge panel, which includes the assessment of language correctness.

To assess the reliability of the LLM judge panel, we compare its evaluations with those provided by human annotators. Specifically, we collect responses from five different LLMs to 300 English questions in the \textit{monolingual retrieval} setting, yielding a total of 1,500 responses. The LLM judge panel is then used to evaluate the correctness of each response against the gold answer. In parallel, we recruit three annotators via Amazon Mechanical Turk\footnote{Given the straightforward nature of the evaluation task, we opted to use Mechanical Turk instead of a professional annotation team.} to independently assess the same set of 1,500 responses. These annotators follow the same evaluation guidelines as those used by the LLM-as-a-Judge (see Figure~\ref{prompt:llmjudge}). The majority vote among the three annotators is taken as the final human judgment. To quantify the level of agreement between the LLM judge panel and the human evaluators, we compute Cohen’s kappa, which yields a score of 0.71—indicating substantial agreement between the two evaluation approaches.

\begin{figure}[t]
\centering
\includegraphics[scale=0.34,trim=0 0 0 0]{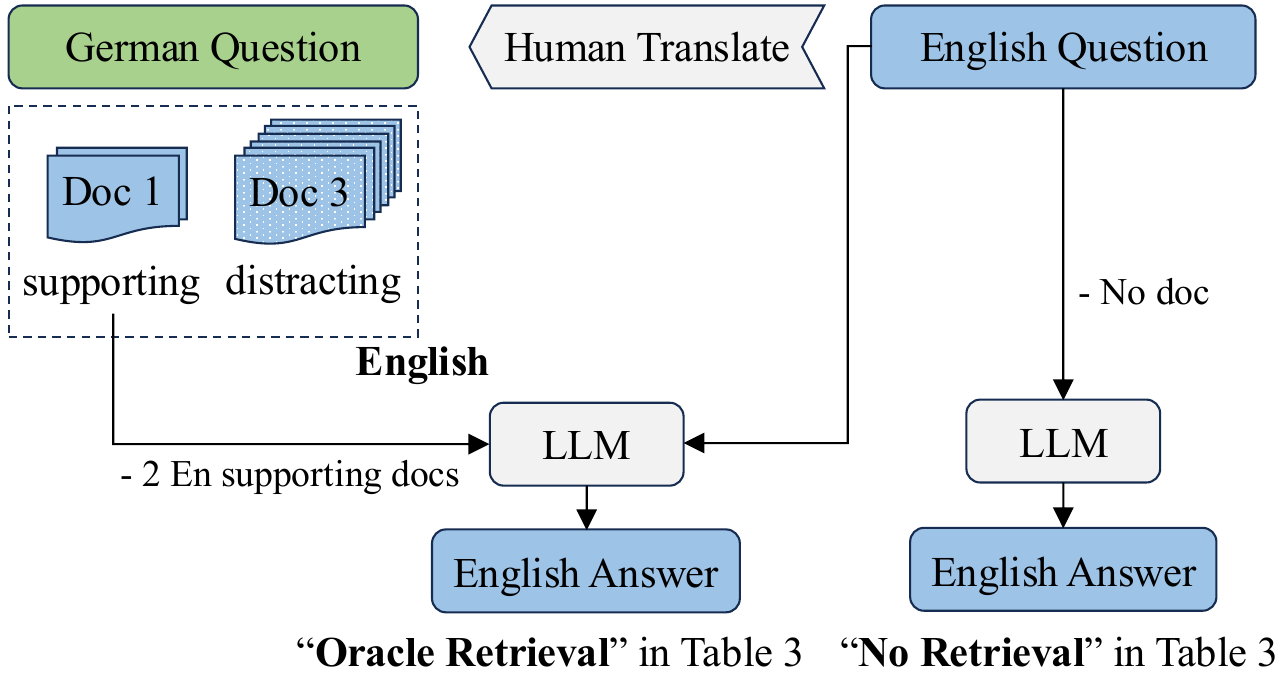}
\setlength{\abovecaptionskip}{6pt}
\setlength{\belowcaptionskip}{-16pt}
\caption{Experimental settings in Table \ref{table:property}.}
\label{fig:corpus_analysis}
\end{figure}

\begin{figure*}[t]
\centering
\includegraphics[scale=0.46,trim=0 0 0 0]{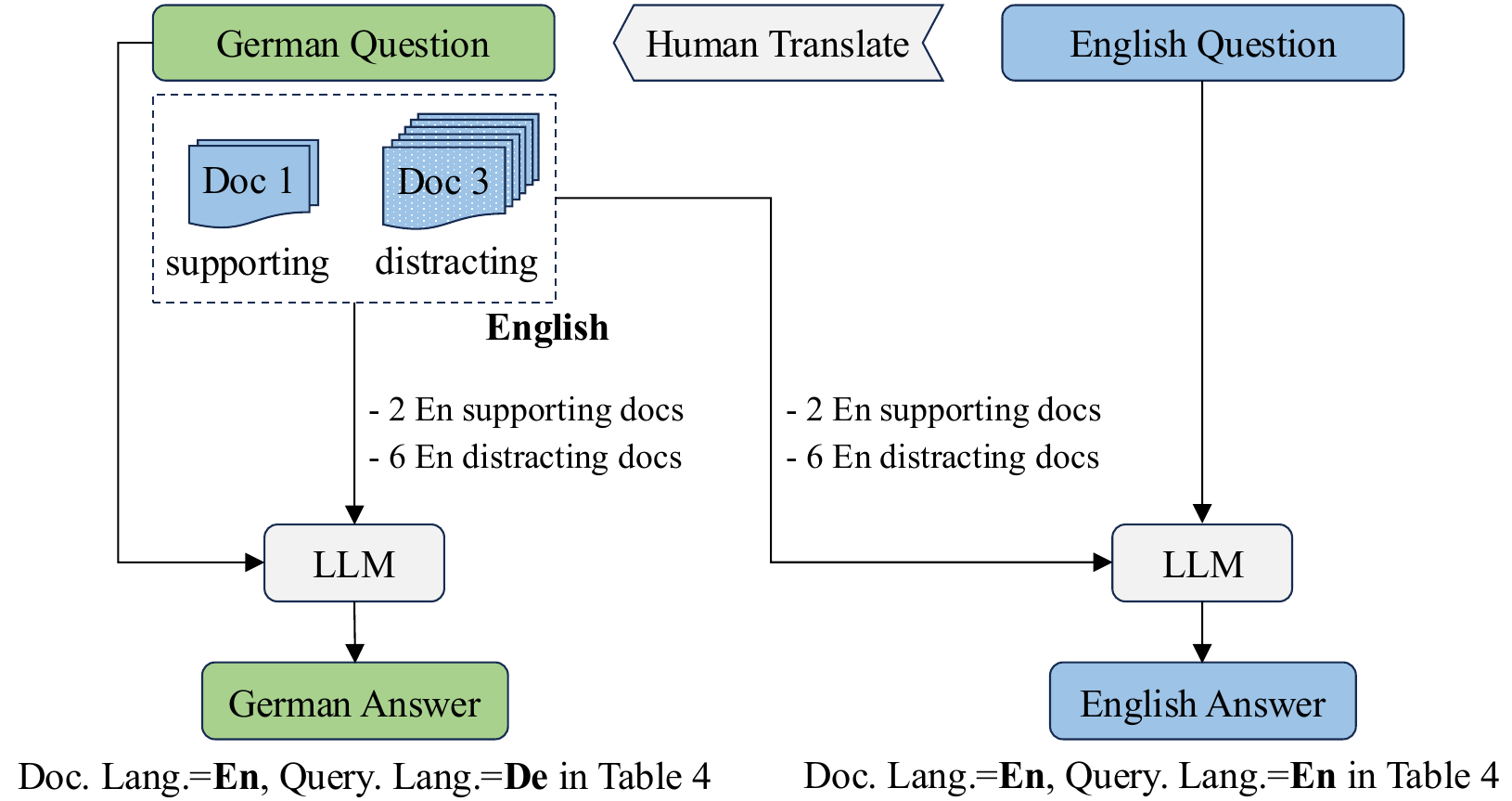}
\setlength{\abovecaptionskip}{6pt}
\setlength{\belowcaptionskip}{6pt}
\caption{Experimental settings in Table \ref{table:mono}. Here, we use English (\textbf{En}) and German (\textbf{De}) as examples.}
\label{fig:mono_eval}
\end{figure*}

\begin{figure*}[t]
\centering
\includegraphics[scale=0.46,trim=0 0 0 0]{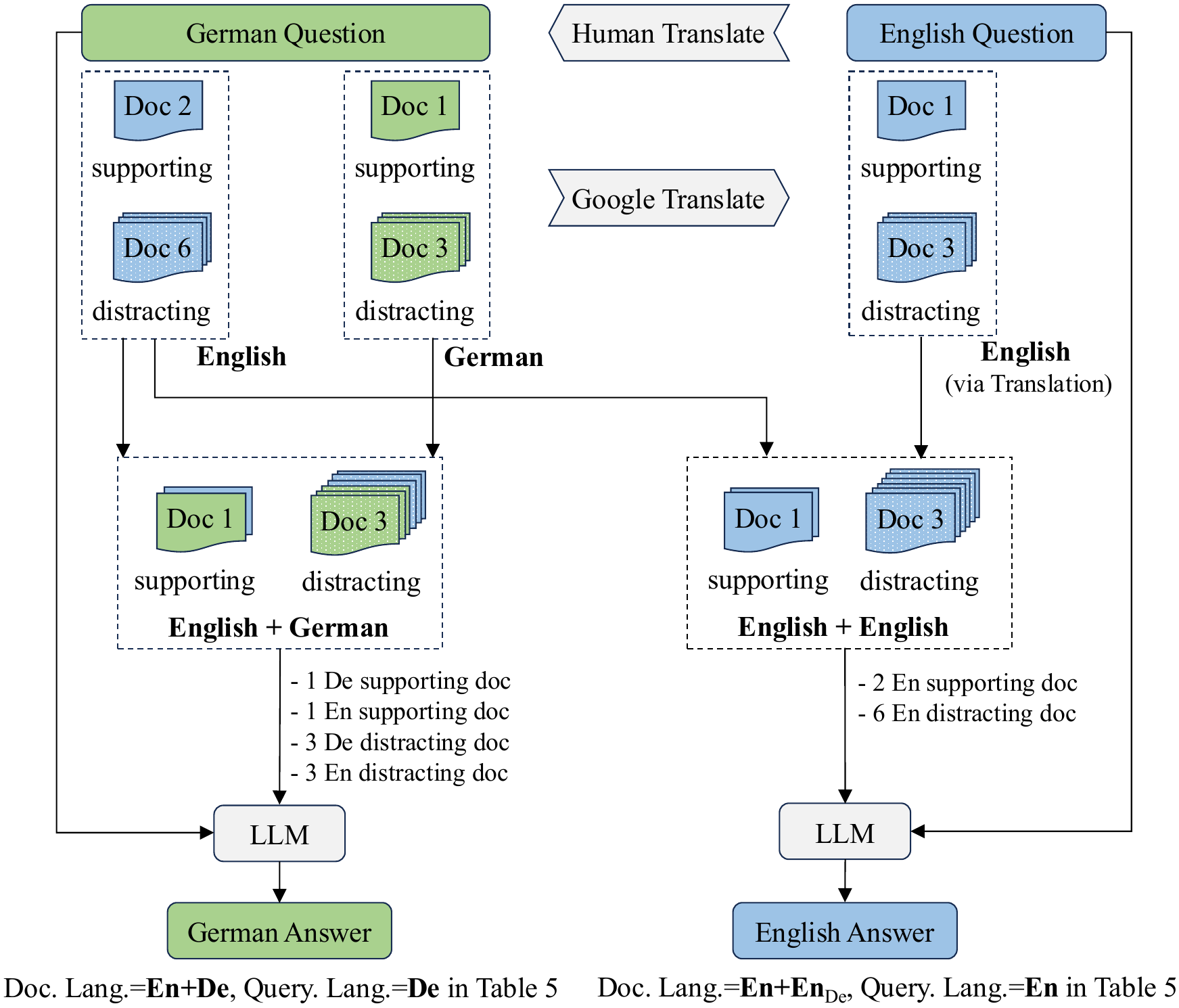}
\setlength{\abovecaptionskip}{6pt}
\setlength{\belowcaptionskip}{6pt}
\caption{Experimental settings in Table \ref{table:multi}. Here, we use English + German (\textbf{En+De}) as an example.}
\label{fig:multi_eval}
\end{figure*}

\begin{figure*}[t]
\centering
\includegraphics[scale=0.40,trim=0 0 0 0]{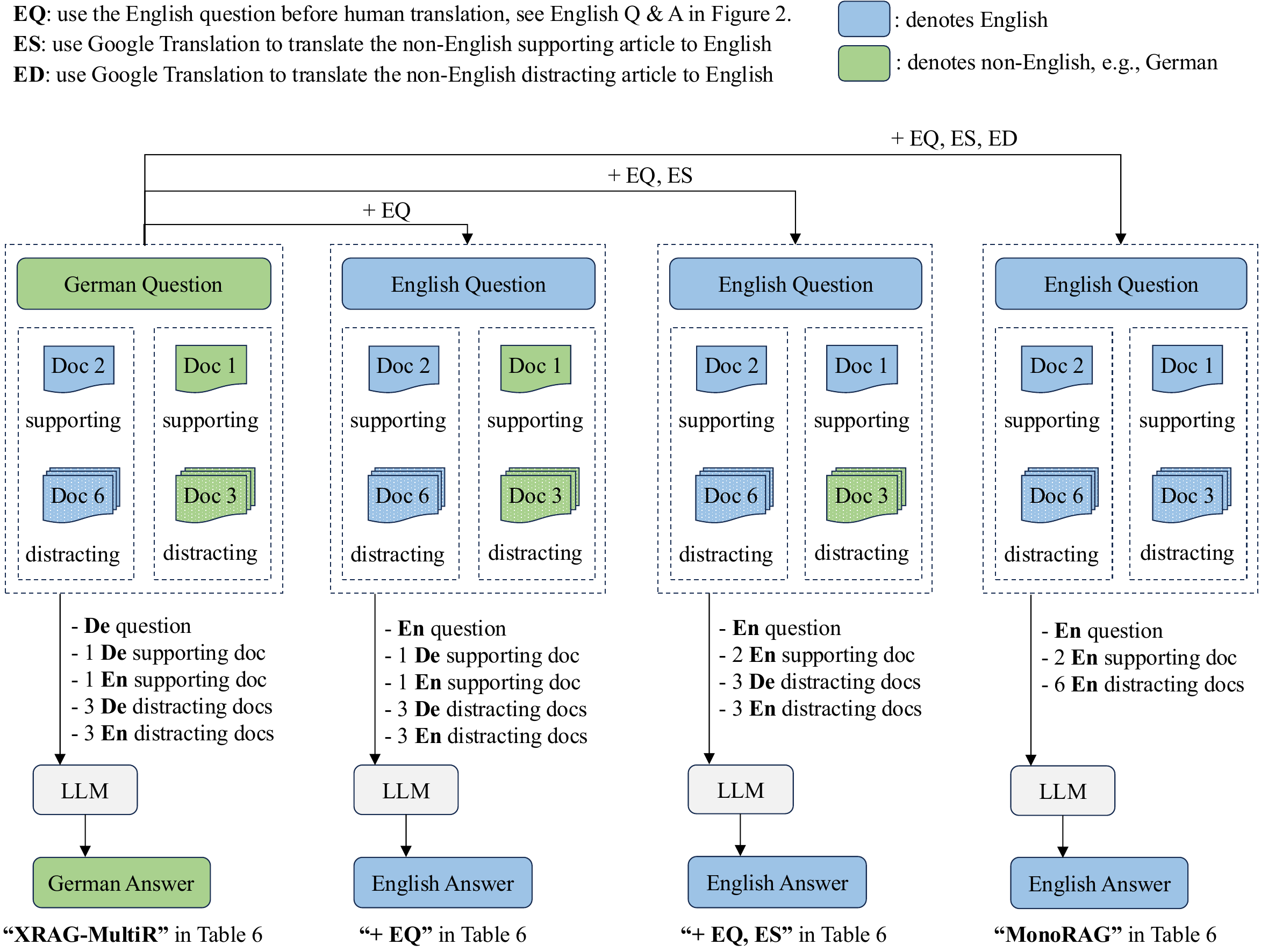}
\setlength{\abovecaptionskip}{6pt}
\setlength{\belowcaptionskip}{6pt}
\caption{Experimental settings in Table \ref{table:control}. Here, we use English + German (\textbf{En+De}) as an example.}
\label{fig:controlled_analysis}
\end{figure*}

\vspace{-2pt}
\subsection{Settings in Different Tables}
\vspace{-2pt}
To facilitate the interpretation of the results presented in different tables, we provide diagrams illustrating the corresponding experimental setups. Specifically, Figures \ref{fig:corpus_analysis}, \ref{fig:mono_eval}, \ref{fig:multi_eval}, and \ref{fig:controlled_analysis} depict the experimental configurations associated with Tables \ref{table:property}, \ref{table:mono}, \ref{table:multi}, and \ref{table:control}, respectively.

\begin{table}[!t]
\centering
\Large
\scalebox{0.595}{
\begin{tabular}{l|cccc|c}
\toprule
\textbf{Claude 3.5} & \hspace{0.0em}\textbf{En+De}\hspace{0.0em} & \hspace{0.0em}\textbf{En+Es}\hspace{0.0em} & \hspace{0.0em}\textbf{En+Zh}\hspace{0.0em} & \hspace{0.0em}\textbf{En+Ar} & \textbf{Avg.} \\ \midrule
XRAG-MultiR       & 45.67          & 42.67           & 48.00           & 39.67  &   44.00  \\
+EQ           & 49.00          & 40.67           & 45.33           & 42.67    & 44.42      \\
+EQ, ES       & 46.67          & 44.00           & 46.33           & 47.67 & 46.17 \\
MonoRAG   & 51.00          & 46.33          & 46.67          & 47.33  & 47.83         \\ \bottomrule
\end{tabular}}
\setlength{\abovecaptionskip}{6pt}
\setlength{\belowcaptionskip}{2pt}
\caption{Controlled analysis of Claude Sonnet 3.5 on the \textit{multilingual retrieval} setting of XRAG, replacing questions (EQ), supporting articles (ES), and distracting articles (ED) with their English counterparts from the English monolingual RAG settings (see Figure \ref{fig:controlled_analysis}). "XRAG-MultiR" is the \textit{multilingual retrieval} setting, and "MonoRAG" (+EQ, +ES, +ED) is the English monolingual RAG baseline setting.}
\label{table:control_claude}
\end{table}

\begin{table}[!t]
\centering
\Large
\scalebox{0.595}{
\begin{tabular}{l|cccc|c}
\toprule
\textbf{Mistral-large} & \hspace{0.0em}\textbf{En+De}\hspace{0.0em} & \hspace{0.0em}\textbf{En+Es}\hspace{0.0em} & \hspace{0.0em}\textbf{En+Zh}\hspace{0.0em} & \hspace{0.0em}\textbf{En+Ar} & \textbf{Avg.} \\ \midrule
XRAG-MultiR       & 42.00          & 39.33           & 37.33           & 32.00  &   37.67  \\
+EQ           & 42.67          & 34.67           & 40.67           & 38.00    & 39.00      \\
+EQ, ES       & 43.33          & 40.00           & 47.33           & 45.33 & 44.00 \\
MonoRAG   & 45.67          & 43.00          & 48.67          & 43.67  & 45.25         \\ \bottomrule
\end{tabular}}
\setlength{\abovecaptionskip}{6pt}
\setlength{\belowcaptionskip}{2pt}
\caption{Controlled analysis of Mistral-large on the \textit{multilingual retrieval} setting of XRAG, replacing questions (EQ), supporting articles (ES), and distracting articles (ED) with their English counterparts from the English monolingual RAG settings (see Figure \ref{fig:controlled_analysis}).}
\label{table:control_mistral}
\end{table}

\begin{table}[!t]
\centering
\Large
\scalebox{0.595}{
\begin{tabular}{l|cccc|c}
\toprule
\textbf{Command-R+} & \hspace{0.0em}\textbf{En+De}\hspace{0.0em} & \hspace{0.0em}\textbf{En+Es}\hspace{0.0em} & \hspace{0.0em}\textbf{En+Zh}\hspace{0.0em} & \hspace{0.0em}\textbf{En+Ar} & \textbf{Avg.} \\ \midrule
XRAG-MultiR       & 40.00          & 40.33           & 36.33           & 32.00  &   37.17  \\
+EQ           & 41.00          & 34.00           & 40.67           & 40.33    & 39.00      \\
+EQ, ES       & 43.33          & 36.00           & 50.33           & 43.33 & 43.25 \\
MonoRAG   & 43.67          & 42.33          & 49.67          & 43.33  & 44.75         \\ \bottomrule
\end{tabular}}
\setlength{\abovecaptionskip}{6pt}
\setlength{\belowcaptionskip}{-8pt}
\caption{Controlled analysis of Command-R+ on the \textit{multilingual retrieval} setting of XRAG, replacing questions (EQ), supporting articles (ES), and distracting articles (ED) with their English counterparts from the English monolingual RAG settings (see Figure \ref{fig:controlled_analysis}).}
\label{table:control_command}
\end{table}

\vspace{-4pt}
\section{More Experimental Results}
\vspace{-2pt}
\subsection{Controlled Analysis}
\vspace{-2pt}
\label{app:control}
In Section \ref{sec:multi_res}, we perform a controlled analysis on the language of query, distracting articles, and supporting articles in the cross-lingual RAG with \textit{multilingual retrieval}, using GPT-4o as the primary model. Here, we extend the analysis to additional LLMs: results for Claude 3.5 Sonnet are shown in Table \ref{table:control_claude}, Mistral-large in Table \ref{table:control_mistral}, and Command-R+ in Table \ref{table:control_command}. Similar results are observed across these LLMs: translating the supporting articles from non-English to English leads to the most improvement (i.e., +ES). This suggests that the primary challenge does not appear to lie in non-English text generation but rather in reasoning over retrieved information across languages.

\end{document}